\documentclass{article}

\usepackage{microtype}
\usepackage{graphicx}
\usepackage{subfigure}
\usepackage{booktabs} % for professional tables
\usepackage{multirow}
\usepackage{mathtools}
\usepackage{amsmath,amssymb,amsthm,amsfonts}
\usepackage{xcolor}

\usepackage{mdframed}   % for framing

\usepackage{hyperref}

\usepackage[export]{adjustbox}

\usepackage[accepted]{icml2019}

\icmltitlerunning{Fairwashing: the risk of rationalization}

\begin{document}

\newmdtheoremenv[%
linecolor=gray,leftmargin=0,%
rightmargin=0,
%backgroundcolor=gray!20,%
backgroundcolor=white,%
innertopmargin=0pt,%
ntheorem]{myExample}{RuleList}[section]

\def\model#1#2#3{#1 \colon  #2 \to #3}

\def\rationSet{\mathcal{T}}

\def\rll{(d_p, \delta_p, q_0, K)}

\def\rlll{(d'_p, \delta'_p, q'_0, K')}

\def\rl{d=(d_p, \delta_p, q_0, K)}

\def\obj{R(\cdot)=\mathsf{misc}(\cdot) + \lambda K}
\def\misc{misc(\cdot)}

\def\objOld{R(d,x,y)=\mathsf{misc}(d,x,y) + \lambda K}
\def\miscOld{misc(d,x,y)}

\def\objj{\mathsf{obj}(\cdot)=(1-\beta)\mathsf{misc}(\cdot) + \beta \mathsf{unfairness}(\cdot) + \lambda K}

\def\bbox{\mathcal{B}}

\def\sg{X_{S_g}}

\def\group{suing group}

\def\badml{\texttt{LaundryML}}
\def\badmll{\texttt{LaundryML-local}}
\def\badmlg{\texttt{LaundryML-global}}

\newtheorem{definition}[theorem]{Definition} 

%macros to update the paper
\def\corr#1{\textcolor{red}{#1}}

\twocolumn[
\icmltitle{Fairwashing: the risk of rationalization}

\begin{icmlauthorlist}
\icmlauthor{Ulrich A{\"\i}vodji}{uqam}
\icmlauthor{Hiromi Arai}{riken,jst}
\icmlauthor{Olivier Fortineau}{paristech}
\icmlauthor{S{\'e}bastien Gambs}{uqam}
\icmlauthor{Satoshi Hara}{osaka}
\icmlauthor{Alain Tapp}{udem,mila}
\end{icmlauthorlist}

\icmlaffiliation{uqam}{Universit{\'e} du Qu{\'e}bec {\`a} Montr{\'e}al}
\icmlaffiliation{riken}{RIKEN Center for Advanced Intelligence Project}
\icmlaffiliation{jst}{JST PRESTO}
\icmlaffiliation{paristech}{ENSTA ParisTech}
\icmlaffiliation{osaka}{Osaka University}
\icmlaffiliation{udem}{UdeM}
\icmlaffiliation{mila}{MILA}

\icmlcorrespondingauthor{Ulrich A{\"\i}vodji}{aivodji.ulrich@courrier.uqam.ca}

\icmlkeywords{Machine Learning, Fairness, Interpretability}

\vskip 0.3in
]

\printAffiliationsAndNotice{}  % leave blank if no need to mention equal contribution
%\printAffiliationsAndNotice{\icmlEqualContribution}

\begin{abstract}
Black-box explanation is the problem of explaining how a machine learning model -- whose internal logic is hidden to the auditor and generally complex -- produces its outcomes. 
Current approaches for solving this problem include model explanation, outcome explanation as well as model inspection. 
While these techniques can be beneficial by providing interpretability, they can be used in a negative manner to perform fairwashing, which we define as promoting the false perception that a machine learning model respects some ethical values. 
In particular, we demonstrate that it is possible to systematically rationalize decisions taken by an unfair black-box model using the model explanation as well as the outcome explanation approaches with a given fairness metric. 
Our solution, LaundryML, is based on a regularized rule list enumeration algorithm whose objective is to search for fair rule lists approximating an unfair black-box model. 
We empirically evaluate our rationalization technique on black-box models trained on real-world datasets and show that one can obtain rule lists with high fidelity to the black-box model while being considerably less unfair at the same time.
\end{abstract}

\section{Introduction}
In recent years, the widespread use of machine learning models in high stakes decision-making systems (\emph{e.g.}, credit scoring~\cite{siddiqi2012credit}, predictive justice~\cite{kleinberg2017human} or medical diagnosis~\cite{caruana2015intelligible}) combined with proven risks of incorrect decisions -- such as people being wrongly denied parole~\cite{wexler2017computer} -- has considerably raised the public demand for being able to provide explanations to algorithmic decisions. 

In particular, to ensure transparency and explainability in algorithmic decision processes, several initiatives have emerged for regulating the use of machine learning models. 
For instance, in Europe, the new General Data Protection Regulation has a provision requiring explanations for the decisions of machine learning models that have a significant impact on individuals~\cite{goodman2016european}. 

Existing methods to achieve explainability include transparent-box design and black-box explanation (also called post-hoc explanation)~\cite{lipton2016mythos,lepri2017fair,montavon2018methods,guidotti2018survey}.
The former consists in building transparent models, which are inherently interpretable by design. 
This approach requires the cooperation of the entity responsible for the training and usage of the model.
In contrast, the latter involves an adversarial setting in which the black-box model, whose internal logic is hidden to the auditor, is reversed-engineered to create an interpretable surrogate model. 

While many recent works have focused on black-box explanations, their use can be problematic in high stakes decision-making systems~\cite{reddix2011credit,wexler2017tradesecret} as explanations can be unreliable and misleading ~\cite{rudin2018please,melis2018towards}. 
Current techniques for providing black-box explanations include model explanation, outcome explanation and model inspection~\cite{guidotti2018survey}. 
Model explanation consists in building an interpretable model to explain the whole logic of the black box while outcome explanation only cares about providing a local explanation of a specific decision. 
Finally, model inspection consists of all the techniques that can help to understand (\emph{e.g.}, through visualizations or quantitative arguments) the influence of the attributes of the input on the black-box decision.

Since the right to explanation as defined in current regulations~\cite{goodman2016european} does not give precise directives on what it means to provide a ``valid explanation''~\cite{wachter2017right,edwards2017slave}, there is a legal loophole that can be used by dishonest companies to cover up the possible unfairness of their black-box models by providing misleading explanations. 
In particular, due to the growing importance of the concepts of fairness in machine learning~\cite{barocas2016big}, a company might be tempted to perform \emph{fairwashing}, which we define as promoting the false perception that the learning models used by the company are fair while it might not be so.

In this paper, our main objective is to raise the awareness of this issue, which we believe to be a serious concern related to the use of black-box explanation. 
In addition, to provide concrete evidence on this possible risk, we show that it is possible to forge a fairer explanation from a truly unfair black box through a process that we coin as \emph{rationalization}. 

In particular, we propose a systematic rationalization technique that, given black-box access to a model $f$, produces an ensemble $C$ of interpretable models $c \approx f$ that are fairer than the black-box according to a predefined fairness metric. 
From this set of plausible explanations, a dishonest entity can pick a model to achieve fairwashing. 
One of the strength of our approach is that it is agnostic to both the black-box model and the fairness metric. 
To demonstrate the genericity of our technique, we show that the rationalization can be used to explain globally the decision of a black-box model (\emph{i.e.}, rationalization of model explanation) as well as its decision for a particular input (\emph{i.e.}, rationalization of outcome explanation). 
While our approach is mainly presented in this paper as a proof-of-concept to raise awareness on this issue, we believe that the number of ways rationalization can be instantiated are endless.

The outline of the paper is as follows. 
First in Section \ref{sec:background}, we review the related work on interpretability and fairness necessary to the understanding of our work. 
Afterwards in Section \ref{sec:solution}, we formalize the rationalization problem before introducing \badml, our algorithm for the enumeration of rule lists that can be used to perform fairwashing.
Finally, in Section \ref{sec:experiments} we report on the evaluation obtained when applying on rationalization approach on two real datasets before concluding in Section \ref{sec:conclusion}.

\section{Related work}
\label{sec:background}

\subsection{Interpretability and explanation}
\label{interp-exp}

In the context of machine learning, interpretability can be defined as the ability to explain or to provide the meaning in understandable terms to a human~\cite{doshi2017towards}.
An explanation is an interface between humans and a decision process that is both an accurate proxy of the decision process and understandable by humans~\cite{guidotti2018survey}. 
Examples of such interfaces include linear models~\cite{ribeiro2016should}, decision trees~\cite{breiman2017classification}, rule lists~\cite{rivest1987learning,letham2015interpretable,angelino2018learning} and rule sets~\cite{li2002mining}.
In this paper, we focus on two black-box explanation tasks, namely model explanation and outcome explanation~\cite{guidotti2018survey}.

\begin{definition}[Model explanation]
\label{def:model-exp}
\emph{Given a black-box model $b$ and a set of instances $X$, the model explanation problem consists in finding an explanation $E \in \mathcal{E}$ belonging to a human-interpretable domain $\mathcal{E}$, through an interpretable global predictor $c_g = f(b,X)$ derived from the black-box $b$ and the instances $X$ using some process $f(\cdot, \cdot)$.}
\end{definition}

For example, if we choose the domain $\mathcal{E}$ to be the set of decision trees, model explanation amounts to identifying a decision tree that approximates well the black-box model.
The identified decision tree can be interpreted as an explanation of the black-box model~\cite{craven1996extracting}.

\begin{definition}[Outcome explanation]
\label{def:outcome-exp}
\emph{Given a black-box model $b$ and an instance $x$, the outcome explanation problem consists in finding an explanation $e \in \mathcal{E}$, belonging to a human-interpretable domain $\mathcal{E}$, through an interpretable local predictor $c_l = f(b,x)$ derived from the black-box $b$ and the instance $x$ using some process $f(\cdot, \cdot)$.}
\end{definition}

For example, choosing the domain $\mathcal{E}$ to be the set of linear models, the outcome explanation amounts to identifying a linear model approximating well the black-box model in the neighbourhood of $x$. 
The identified linear model can be interpreted as a local explanation of the black-box on the instance $x$. 
Approaches such as LIME~\cite{ribeiro2016should} and SHAP~\cite{lundberg2017unified} belong to this class.

\subsection{Fairness in machine learning}
\label{subsec:fairness}
To explain the fairness of machine models, various metrics have been proposed in recent years~\cite{narayanan2018translation,berk2018fairness}.
Roughly there are two distinct families of fairness definitions: \emph{group fairness} and \emph{individual fairness}, which requires decisions to be consistent for individuals that are similar~\cite{dwork2012fairness}.
In this work, we focus on group fairness that requires the approximate equalization of a particular statistical property across groups defined according to the value of a sensitive attribute.
In particular, if the sensitive attribute is binary, its value splits the population into a minority group and a majority group.

One of the common group fairness metrics is \emph{demographic parity}~\cite{calders2009building}, which is defined as the equality of distribution of decisions conditioned to the sensitive attribute.
\emph{Equalized odds}~\cite{hardt2016equality} is another common criterion requiring the false positive and false negative rates be equal across the majority and minority groups.
In a nutshell, demographic parity quantifies biases in both training data and learning, while equalized odds focus on the learning process.
In this paper, we adopt demographic parity to evaluate the overall unfairness in algorithmic decision-making.

Most of the existing tools for quantifying the fairness of a machine learning model do not address the interpretability issue at the same time.
Possible exceptions include \texttt{AI Fairness 360} that provides visualizations of attribute bias and feature bias localization~\cite{bellamy2018ai}, \texttt{DCUBE-GUI}~\cite{berendt2012exploring} that has an interface for visualizing the unfairness scores of association rules resulted from data mining as well as \texttt{FairML}~\cite{adebayo2016fairml} that computes the relative importance of each attribute into the prediction outcome. 
While these tools actually provide a form of model inspection, they cannot be used directly for model or outcome explanations.

\subsection{Rule Lists}
A rule list~\cite{rivest1987learning,letham2015interpretable,angelino2018learning} $\rl{}$ of length $K \geq 0$ is a $(K+1)-$tuple consisting of $K$ distinct association rules $r_k = p_k \to q_k$, in which $p_k \in d_p$ is the antecedent of the association rule and $q_k \in \delta_p$ its corresponding consequent, followed by a default prediction $q_0$. 
An equivalent way to represent a rule list is in the form of a decision tree whose shape is a comb.
Making a prediction using a rule list means that its rules are applied sequentially until one rule triggers, in which case the associated outcome is reported. 
If no rules are trigerred, then the default prediction is reported.

\subsection{Optimal rule list and enumeration of rule lists}
\label{subsec:mlEnum}

CORELS~\cite{angelino2018learning} is a supervised machine learning algorithm which takes as input a training set with $n$ predictor variables, all assumed to be binary.
First, it represents the search space of the rule lists as a $n$-level trie whose root node has $n$ children, formed by the $n$ predictor variables, each of which has $n-1$ children, composed of all the predictor variables except the parent, and so on.
Afterwards, it considers for a rule list $\rl{}$, an objective function defined as a regularized empirical risk: 
\begin{align*}
\obj{},
\end{align*}
in which $\misc{}$ is the misclassification error and $\lambda \geq 0$ the regularization parameter used to penalize longer rule lists. 
Finally, CORELS selects the rule list that minimizes the objective function. 
To realize this, it uses an efficient branch-and-bound algorithm to prune the trie. 
In addition, on high-dimensional datasets (\emph{i.e.}, with a large number of predictor variables), CORELS uses both the number of nodes to be evaluated in the trie and the upper bound on the remaining search space size as a stopping criterion to identify suboptimal rule lists.

In~\cite{hara2018approximate}, the authors propose an algorithm based on the Lawler's framework~\cite{lawler1972procedure}, which allows to successively compute the optimal solution and then construct sub-problems excluding the solution obtained. 
In particular, the authors have combined this framework with CORELS~\cite{angelino2018learning} to compute the exact enumeration for rule lists. 
In a nutshell, this algorithm takes as input the set $T$ of all antecedents.
It maintains a heap $\mathcal{H}$, whose priority is the objective function value of the rule lists and a list $\mathcal{M}$ for the enumerated models. 
First, it starts by computing the optimal rule $m=\mathsf{CORELS}(T) = \rll{}$ for $T$ using the CORELS algorithm. 
Afterward, it inserts the tuple $(m, T, \emptyset)$ into the heap. 
Finally, the algorithm repeats the following steps until the stopping criterion is met:
\begin{itemize}
\item (Step 1) Extract the tuple $(m, S, F)$ with the maximum priority value from $\mathcal{H}$.
\item (Step 2) Output $m$ as the $i-$th model if $m \notin \mathcal{M}$.
\item (Step 3) Branch the search space: compute and insert $m' =\mathsf{CORELS}(S \setminus \{t_j\})$ into $\mathcal{H}$ for all $t_j \in \delta_p$.
\end{itemize}
%\\ (Step 1) Extract the tuple $(m, S, F)$ with the maximum priority value from $\mathcal{H}$.\\ (Step 2) Output $m$ as the $i-$th model if $m \notin \mathcal{M}$.\\ (Step 3) Branch the search space: compute and insert $m' =\mathsf{CORELS}(S \setminus \{t_j\})$ into $\mathcal{H}$ for all $t_j \in \delta_p$. 

\section{Rationalization}
\label{sec:solution}

In this section, we first formalize the problem of the rationalization before introducing \badml, our regularized enumeration algorithm that can be used to perform fairwashing with respect to black-box machine learning models.

\subsection{Problem formulation}
\label{subsec:pblformulation}

A typical scenario that we envision is the situation in which a company wants to perform fairwashing because it is afraid of a possible audit evaluating the fairness of the machine learning it uses to personalize the services provided to its users. 
In this context, rationalization consists in finding an interpretable surrogate model $c$ approximating a black-box model $b$, such that $c$ is fairer than $b$. 
To achieve fairwashing, the surrogate model obtained through rationalization could be shown to the auditor (\emph{e.g.}, an external dedicated entity or the users themselves) to convince him that the company is ``clean''.
We distinguish two types of rationalization problems, namely the model rationalization problem and the outcome rationalization problem, which we define hereafter.

\begin{definition}[Model rationalization]
\label{def:model-rationalization}
\emph{Given a black-box model $b$, a set of instances $X$ and a sensitive attribute $s$, the model rationalization problem consists in finding an explanation $E \in \mathcal{E}$ belonging to a human-interpretable domain $\mathcal{E}$, through an interpretable global predictor $c_g = f(b,X)$ derived from the black-box $b$ and the instances $X$ using some process $f(\cdot, \cdot)$, such that $\epsilon(c_g, X, s) > \epsilon(b, X, s)$, for some fairness evaluation metric $\epsilon(\cdot, \cdot, \cdot)$.}
\end{definition}

\begin{definition}[Outcome rationalization]
\label{def:outcome-rationalization}
\emph{Given a black-box model $b$, an instance $x$, a neighborhood $\mathcal{V}(x)$ of $x$  and a sensitive attribute $s$, the outcome rationalization problem consists in finding an explanation $e \in \mathcal{E}$, belonging to a human-interpretable domain $\mathcal{E}$, through an interpretable local predictor $c_l = f(b,x)$ derived from the black-box $b$ and the instance $x$ using some process $f(\cdot, \cdot)$, such that $\epsilon(c_l, \mathcal{V}(x), s) > \epsilon(b, \mathcal{V}(x), s)$, for some fairness evaluation metric $\epsilon(\cdot, \cdot, \cdot)$.}
\end{definition}

In this paper, we will refer to the set of instances for which the rationalization is done as the \group{}.
For the sake of clarity, we also restrict the context to binary attributes and to binary classification. 
However, our approach could also be used to multi-valued attributes.
For instance, multi-valued attributes can be converted to binary ones using a standard approach such as a one-hot encoding. 
For simplicity, we also assume that there is only one sensitive attribute, though our work could also be extended to multiple sensitive attributes.

\subsection{Performance metrics}
\label{sec:rationalization_metrics}

We evaluate the performance of the surrogate by taking into account (1) its fidelity with respect to the original black-box model and (2) its unfairness with respect to a predefined fairness metric. 

For the model rationalization problem, the \emph{fidelity} of the surrogate $c_g$ is defined as its accuracy relative to $b$ on $X$ \cite{craven1996extracting}, which is 
\begin{align*}
\mathsf{fidelity}(c) = \frac{1}{|X|} \sum_{x \in X} \mathbb{I}(c(x)=b(x)).
\end{align*}

For the outcome rationalization problem, the fidelity of the surrogate $c_l$ is $1$ if $c_l(x)=b(x)$ and $0$ otherwise.
 
With respect to fairness, among the numerous existing definitions~\cite{narayanan2018translation}, we rely on the \emph{demographic parity}~\cite{calders2009building} as the fairness metric. Demographic parity requires the prediction to be independent of the sensitive attribute, which is 
\begin{align*}
P(\hat{y}=1 | s= 1) = P(\hat{y}=1 | s= 0),
\end{align*}
in which $\hat{y}$ is the predicted value of $c_g$ for $x \in X$ for the rationalization problem (respectively, the predicted value of $c_l$ for $x \in \mathcal{V}(x)$ for the outcome problem.
Therefore, \emph{unfairness} is defined as 
%\mathsf{unfairness}(c) =
\begin{align*}
|P(\hat{y}=1 | s= 1) - P(\hat{y}=1 | s= 0)|.
\end{align*}

We consider that rationalization leads to \emph{fairwashing} when the fidelity of the interpretable surrogate model is high while at the same time its level of unfairness is significantly lower than the original black-box algorithm.

\subsection{\badml{}: a regularized model enumeration algorithm}

To rationalize unfair decisions, we need to find a good surrogate model that has high fidelity and low unfairness at the same time.
However, finding a single optimal model is difficult in general; indeed it is rarely the case that a single model achieves optimality with respect to two different criteria at the same time.
To bypass this difficulty, we adopt a model enumeration approach.
In a nutshell, our approach works by first enumerating several models with sufficiently high fidelity to the original black-box model and low unfairness.
Afterwards, the approach picks up the model that is convenient for rationalization based on some other criteria or through human inspection.

In this section, we introduce \badml{}, a regularized model enumeration algorithm that enumerates rule lists to achieve rationalization. 
More precisely, we propose two versions of \badml{}, one that solves the model rationalization problem and the other that solves the outcome rationalization.

\badml{} is a modified version of the algorithm presented in Section~\ref{subsec:mlEnum}, which considers both the fairness and fidelity constraints in its search. 
The algorithm is summarized in Algorithm~\ref{alg:mlEnum}.

\begin{algorithm}
\caption{\badml}
\label{alg:mlEnum}
\begin{algorithmic}[1]
\STATE Inputs: $T$, $\lambda$, $\beta$
\STATE Output: $\mathcal{M}$
%\STATEx   \Comment {define the objective function}
\STATE $\mathsf{obj}(\cdot)= (1-\beta)\mathsf{misc}(\cdot) + \beta\mathsf{unfairness}(\cdot) + \lambda K$
\STATE Compute $m = \mathsf{CORELS}(\mathsf{obj},T) = \rll{}$
\STATE Insert $(m, T, \emptyset)$ into the heap
\STATE $\mathcal{M} \gets \emptyset$
\FOR{$i = 1, 2, \ldots$}
    \STATE Extract $(m, S, F)$ from the heap 
     %\STATEx   \Comment {output $m$ as the $i-$th model}
    \IF{$m \notin \mathcal{M}$}
       \STATE $\mathcal{M}\gets \mathcal{M} \cup \{m\}$
    \ENDIF
      %\STATEx   \Comment {terminate when a certain condition is met}
    \IF{$\mathsf{Terminate}(\mathcal{M})= true$}
       \STATE \textbf{break}
    \ENDIF
    %\STATEx   \Comment {branch the search space}
    \FOR{$t_j \in d_p$ and $t_j \notin F$}
    \STATE Compute $m' =\mathsf{CORELS}(\mathsf{obj}, S \setminus \{t_j\} )$
    \STATE Insert $(m', S \setminus \{t_j\}, F )$ into the heap
    \STATE $F \gets F \cup \{t_j\}$
    \ENDFOR
\ENDFOR
\end{algorithmic}
\end{algorithm}

Overall, \badml{} takes as inputs the set of antecedents $T$ and the regularization parameters $\lambda$ and $\beta$, respectively for the rule lists length and the unfairness. 
First, it redefines the objective function for CORELS to penalize both the rule list's length and unfairness (line 3):
\begin{align*}
\objj{},
\end{align*}
in which $\beta \in [0, 1]$  and $\lambda \geq 0$.  
Afterwards, the algorithm runs the enumeration routine described in Section~\ref{subsec:mlEnum} using the new objective function (lines 4--20). 
Using \badml{} to solve the model rationalization problem as well as the outcome rationalization problem is straightforward as described hereafter. 

The model rationalization algorithm depicted in Algorithm~\ref{alg:mlEnum-global} takes as input a black-box access to the black box model $b$, the \group{} $X$ as well as the regularization parameters $\lambda$ and $\beta$. 
First, the members of the \group{} are labeled using the predictions of $b$. 
Afterwards, the labeled dataset $T$ obtained is passed to the \badml{} algorithm. 

\begin{algorithm}
\caption{\badmlg}
\label{alg:mlEnum-global}
\begin{algorithmic}[1]
\STATE Inputs: $X$, $b$, $\lambda$, $\beta$
\STATE Output: $\mathcal{M}$
\STATE $y = b(X)$
\STATE $T = \{X, y \}$
\STATE $\mathcal{M} = \mathsf{\badml(T, \lambda, \beta)}$
\end{algorithmic}
\end{algorithm}

%\begin{algorithm}
%\caption{\badmlg}
%\label{alg:mlEnum-global}
%\begin{algorithmic}[1]
%\STATE Inputs: $\sg{}$, $\bbox{}$, $\mathsf{bin(\cdot)}$, $\lambda$, $\beta$
%\STATE Output: $\mathcal{M}$
%\STATE $y_{S_g} = \bbox{}.\mathsf{predict(\sg{}})$
%\STATE $D_{S_g} = \{\sg{}, y_{S_g} \} $
%\STATE $T = \mathsf{bin(D_{S_g})}$
%\STATE $\mathcal{M} = \mathsf{\badml(T, \lambda, \beta)}$
%\end{algorithmic}
%\end{algorithm}

The rationalized outcome explanation presented in Algorithm~\ref{alg:mlEnum-local} is similar to the rationalized model explanation, but instead of using the labelled dataset, it directly computes for the considered subject $x$ its neighbourhood $T_x =\mathsf{neigh}(x, T_{X})$ and uses it for the model enumeration.

\begin{algorithm}
\caption{\badmll}
\label{alg:mlEnum-local}
\begin{algorithmic}[1]
\STATE Inputs: $x$, $T$, $\mathsf{neigh(\cdot)}$, $\lambda$, $\beta$
\STATE Output: $\mathcal{M}_x$
\STATE $T_x =  \mathsf{neigh(x, T)}$
\STATE $\mathcal{M}_x = \mathsf{\badml(T_x, \lambda, \beta)}$
\end{algorithmic}
\end{algorithm}

In the next section, we report on the results that we have obtained with these two approaches on real-world datasets.

\begin{table*}[t!]
\centering
\caption{Performances of the black-box models. Columns report the accuracy of the black-box models on their respective training set, suing group, and test set, as well as their unfairness on the suing groups.}
\vspace{2mm}
\begin{tabular}{ccccccl}
\toprule
\multirow{2}{*}{\textbf{Dataset}} & \multicolumn{3}{c}{\textbf{Accuracy}} & \multicolumn{3}{c}{\textbf{Unfairness}} \\
 & \textit{Training Set} & \textit{Suing Group} & \textit{Test Set} & \textit{Suing Group} & \multicolumn{2}{c}{\textit{Test Set}} \\
\midrule
Adult Income & 85.22 & 84.31 & 84.9 & 0.13 & \multicolumn{2}{c}{0.18} \\
ProPublica Recidism & 71.42 & 65.86 & 67.61 & 0.17 & \multicolumn{2}{c}{0.13}\\
\bottomrule
\end{tabular}
\vspace{-2mm}
\label{tab:bbox}
\end{table*}

\section{Experimental evaluation}
\label{sec:experiments}

In this section, we describe the experimental setting used to evaluate our rationalization algorithm as well as the results obtained. 
More precisely, we evaluate the unfairness and fidelity of models produced using both model rationalization and outcome rationalization.
The results obtained confirm that it is possible to systematically produce models that are fairer than the black-box models they approximate while still being faithful to the original model.

\subsection{Experimental setting}

\paragraph{Datasets.} We conduct our experiments on two real-world datasets that have been extensively used in the fairness literature due to their biased nature, namely  \textit{Adult Income}~\cite{frank2010uci} and the \textit{ProPublica Recidivism}~\cite{angwin2016machine} datasets.
These datasets have been chosen because they are widely used in the FAT community due to the gender bias (respectively racial bias) in the Adult (respectively the ProPublica Recidivism) dataset.
The Adult Income dataset gathers information about individuals from the 1994 U.S. census, and the objective of the classification task is to predict whether an individual makes more or less than $50,000$\$ per year. 
Overall, this dataset contains $32,561$ instances and $14$ attributes. 
The ProPublica Recidivism dataset gathers criminal offenders records in Florida during 2013 and 2014 with the classification task considered being to predict whether or not a subject will re-offend within two years after the screening. 
Overall, this dataset contains $6,167$ instances and $12$ attributes.

\paragraph{Metrics.} We use both fidelity and unfairness (as defined in Section~\ref{sec:rationalization_metrics}) to evaluate the performance of $\badmlg{}$ and $\badmll{}$. 
In addition to these metrics, we also assess the performance of our approach by auditing both the black-box classifier and the best-enumerated models using \textsf{FairML}~\cite{adebayo2016fairml}. 
In a nutshell, \textsf{FairML} is an automated tool requiring black-box access to a classifier to determine the relative importance of each attribute into the prediction outcome.

\paragraph{Learning the black-box models.} 
We first split each dataset into three subsets, namely the training set, the \group{} and the test set. 
After learning the black-box models on the training sets, we apply them to label the \group{}s, which in turn are used to train the interpretable surrogate models. 
Finally, the test sets are used to evaluate the accuracy and the unfairness of the black-box models. 
In practice for both datasets, we rely on a model trained using \emph{random forest} to play the role of the black-box models.
Table~\ref{tab:bbox} summarizes the performances of the black-box models that have been trained.

\paragraph{Scenarios investigated.} 
As explained in Section~\ref{sec:background}, CORELS requires binary features. 
After the preprocessing, we obtain respectively $28$ and $27$ antecedents for the Adult Income and ProPublica Recidivism dataset. 
In the experiments, we considered the following two scenarios.

\paragraph{(S1) Responding to a \group{}.}
Consider the situation in which a \group{}  complained for unfair decisions and its members request explanations on how the decisions affecting them were made.
In this scenario, a single explanation needs to consistently explain the entire group as much as possible.
Thus, we rationalize the decision using model rationalization based on Algorithm~\ref{alg:mlEnum-global} ($\badmlg{}$).

\paragraph{(S2) Responding to individual claim.}
Imagine that a subject files a complaint following a decision that he deems unfair and that he requests an explanation on how the decision was made.
In this scenario, we only need to respond to each subject independently.
Thus, we rationalize the decision using outcome rationalization based on Algorithm~\ref{alg:mlEnum-local} ($\badmll{}$).

Experiments were conducted on an Intel Core i7 (2.90 GHz, 16GB of RAM). 
The modified version of CORELS is implemented in C++ and based on the original source code of CORELS\footnote{\url{https://goo.gl/qTeUAu}}.
Algorithm~\ref{alg:mlEnum}, Algorithm~\ref{alg:mlEnum-global} as well as Algorithm~\ref{alg:mlEnum-local} were all implemented in Python. 
To implement Algorithm~\ref{alg:mlEnum}, we modified the Lasso enumeration algorithm\footnote{\url{https://goo.gl/CyEU4C}} of~\cite{hara2017enumerate}. 
All our experiments can be reproduced using the code provided in \url{https://github.com/aivodji/LaundryML}.

\begin{figure*}[h!]
   \centering
 \includegraphics[width=0.9\textwidth]{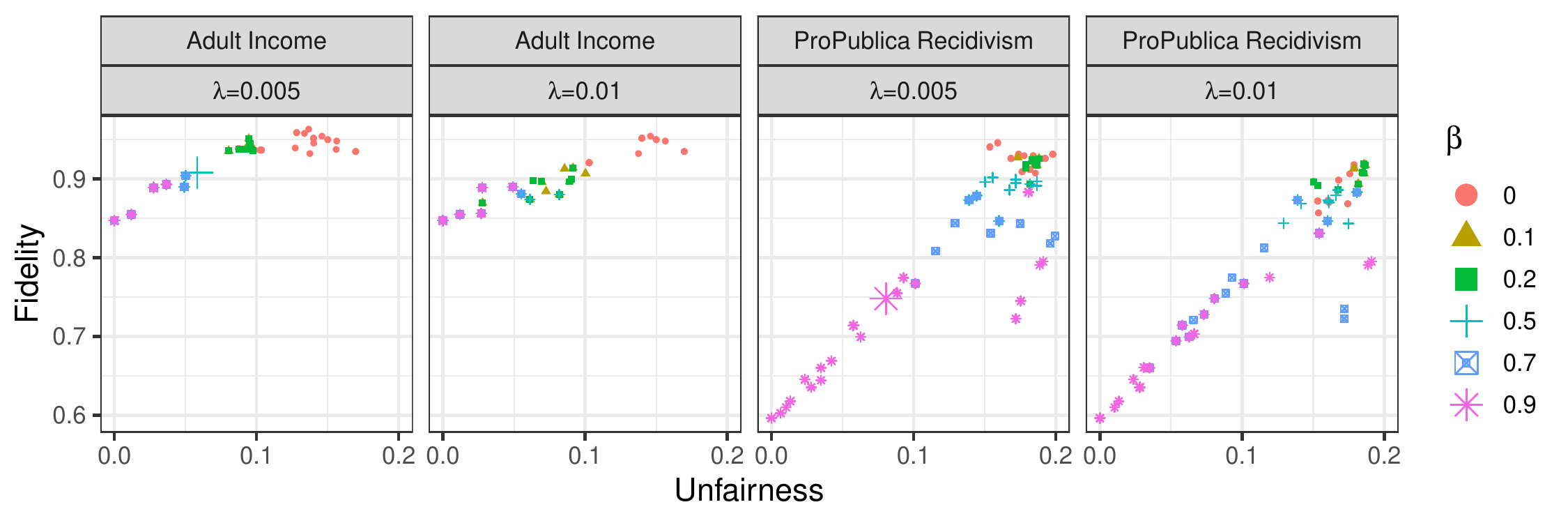}
    \caption{Fidelity and unfairness of rationalized explanation models produced by $\badmlg{}$ on the \group{}s of Adult Income and ProPublica Recidivism datasets.}
    \label{fig:globalExp-train}
\end{figure*}

\begin{figure}[h!]
   \centering
 \includegraphics[width=0.35\textwidth]{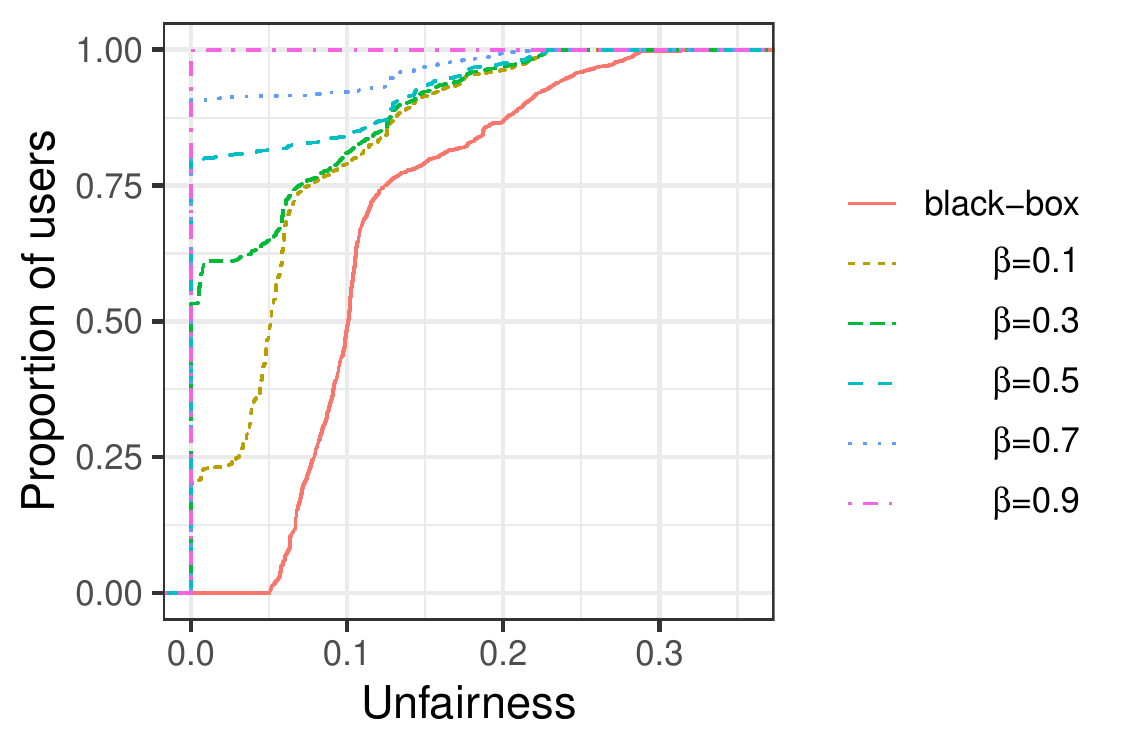}
  \includegraphics[width=0.35\textwidth]{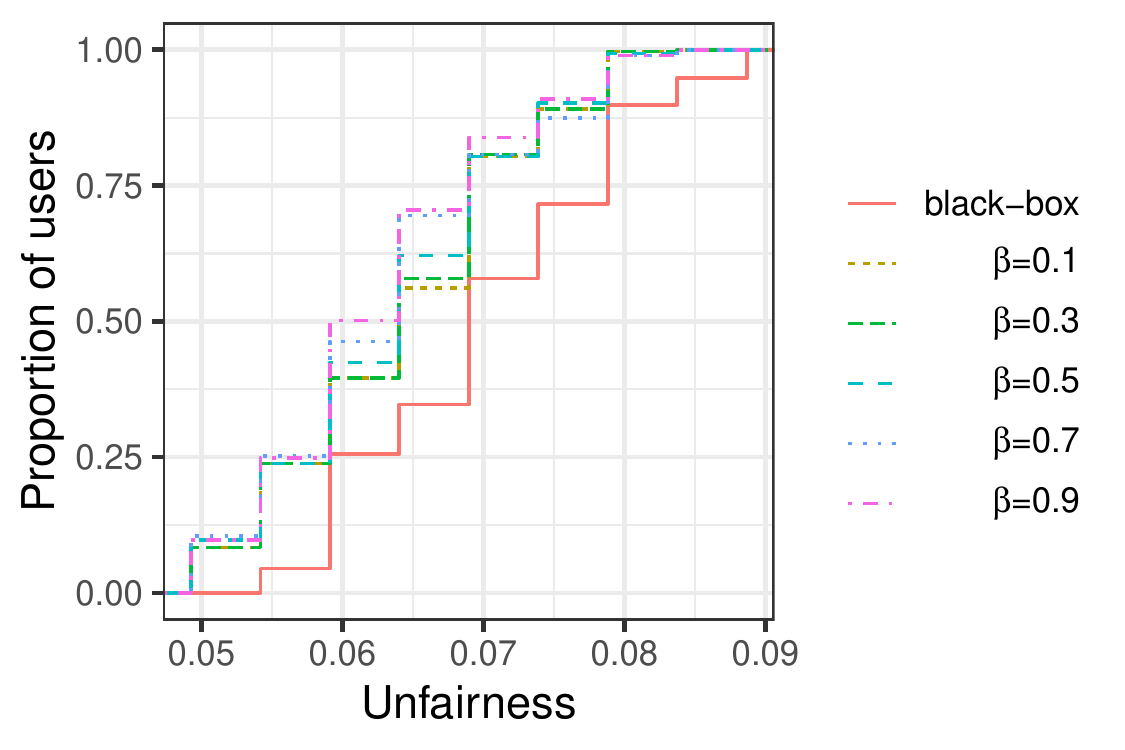}
    \caption{CDFs of the unfairness of the best model found by $\badmll{}$ per user on Adult Income (up) and ProPublica Recidivism (down).}
    \label{fig:local-adult}
\vspace{-2mm}
\end{figure}

\begin{figure*}[htbp]
   \centering
 \includegraphics[width=0.37\textwidth]{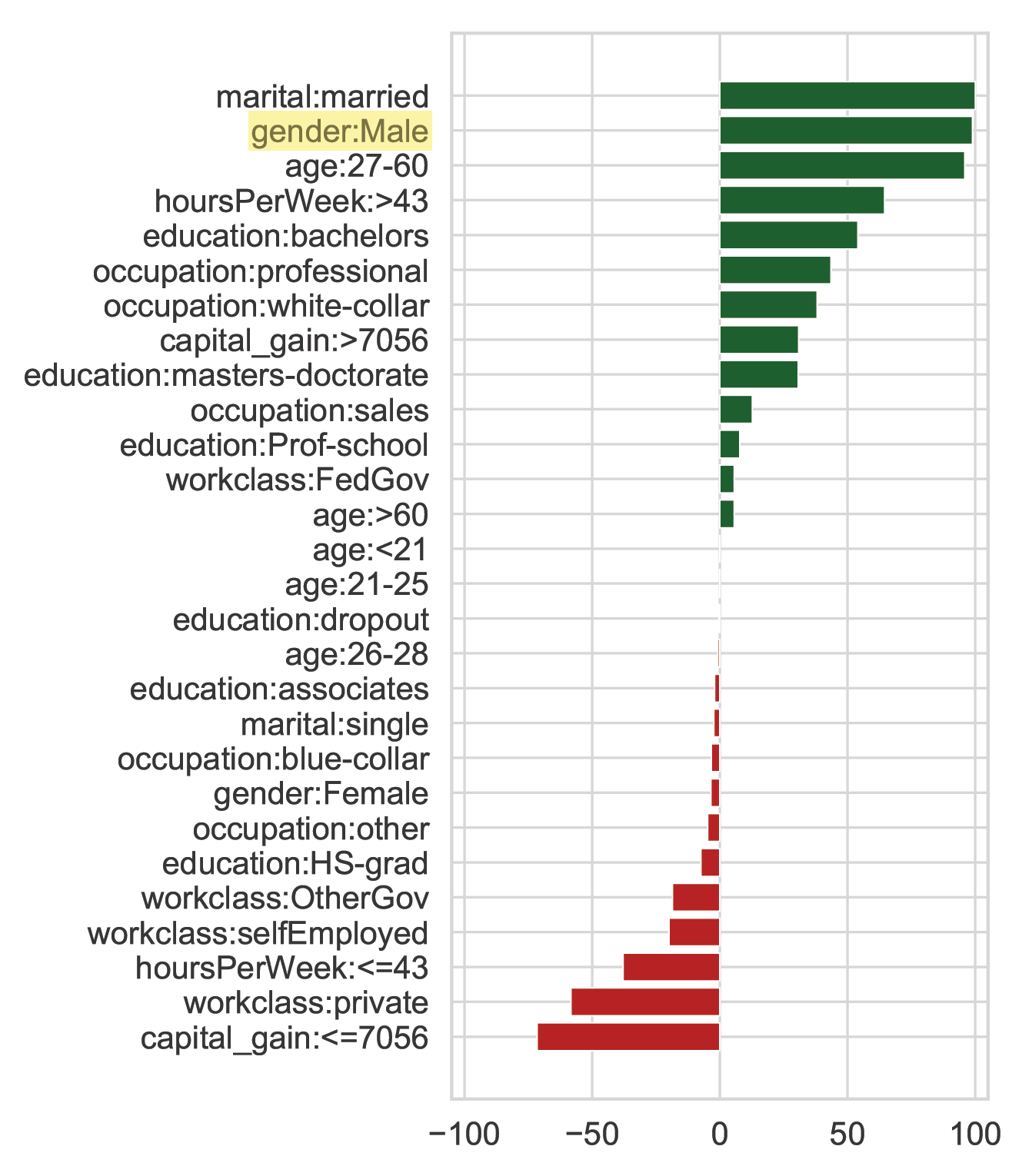}
  \includegraphics[width=0.37\textwidth]{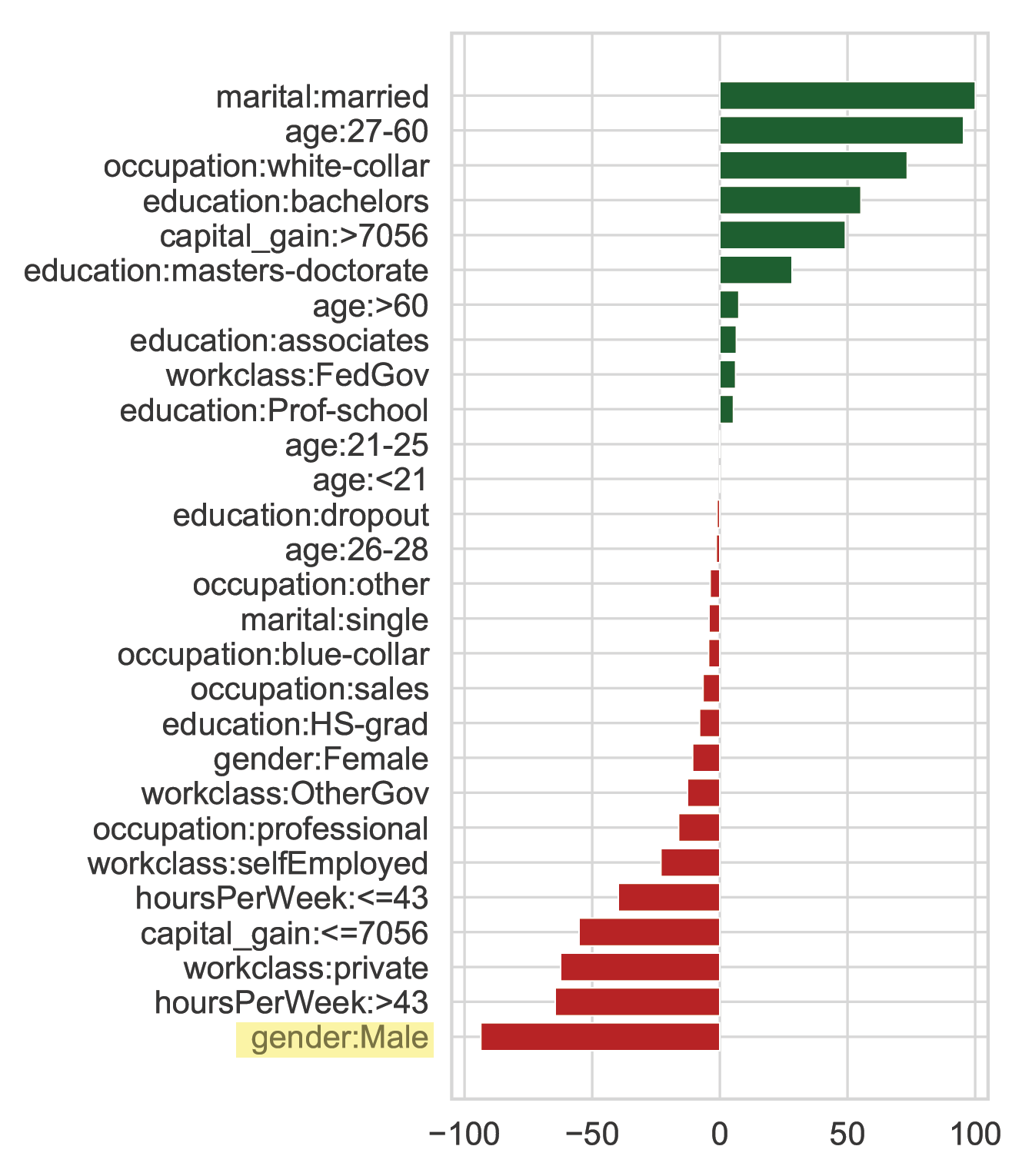}
  \includegraphics[width=0.25\textwidth]{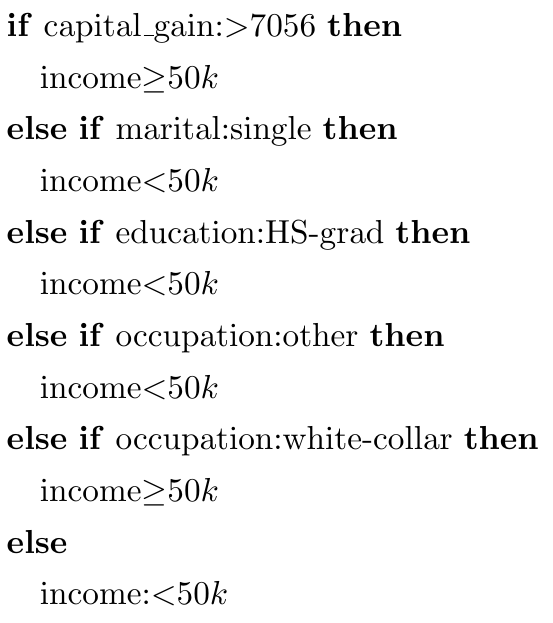}
 \includegraphics[width=0.37\textwidth]{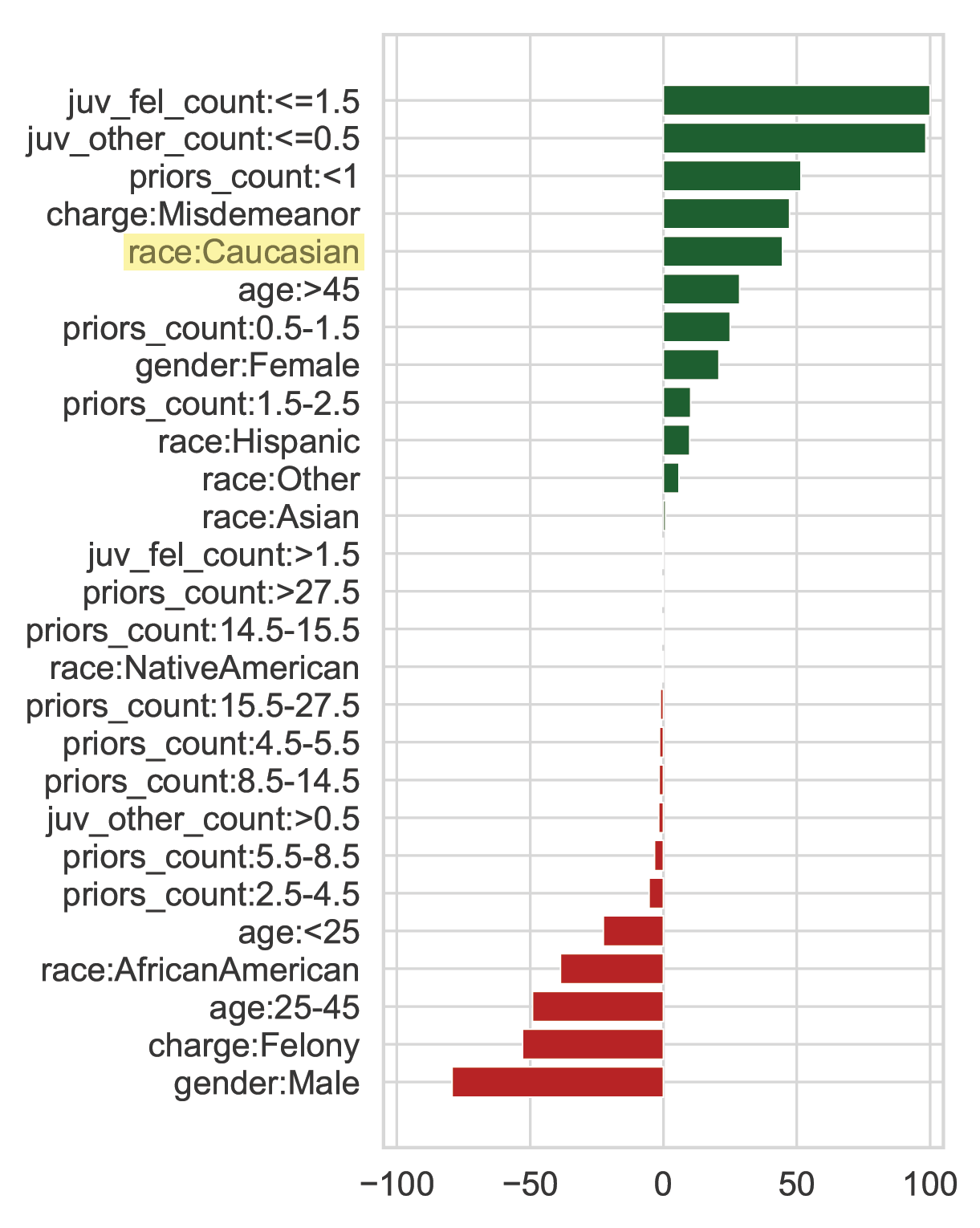}
  \includegraphics[width=0.37\textwidth]{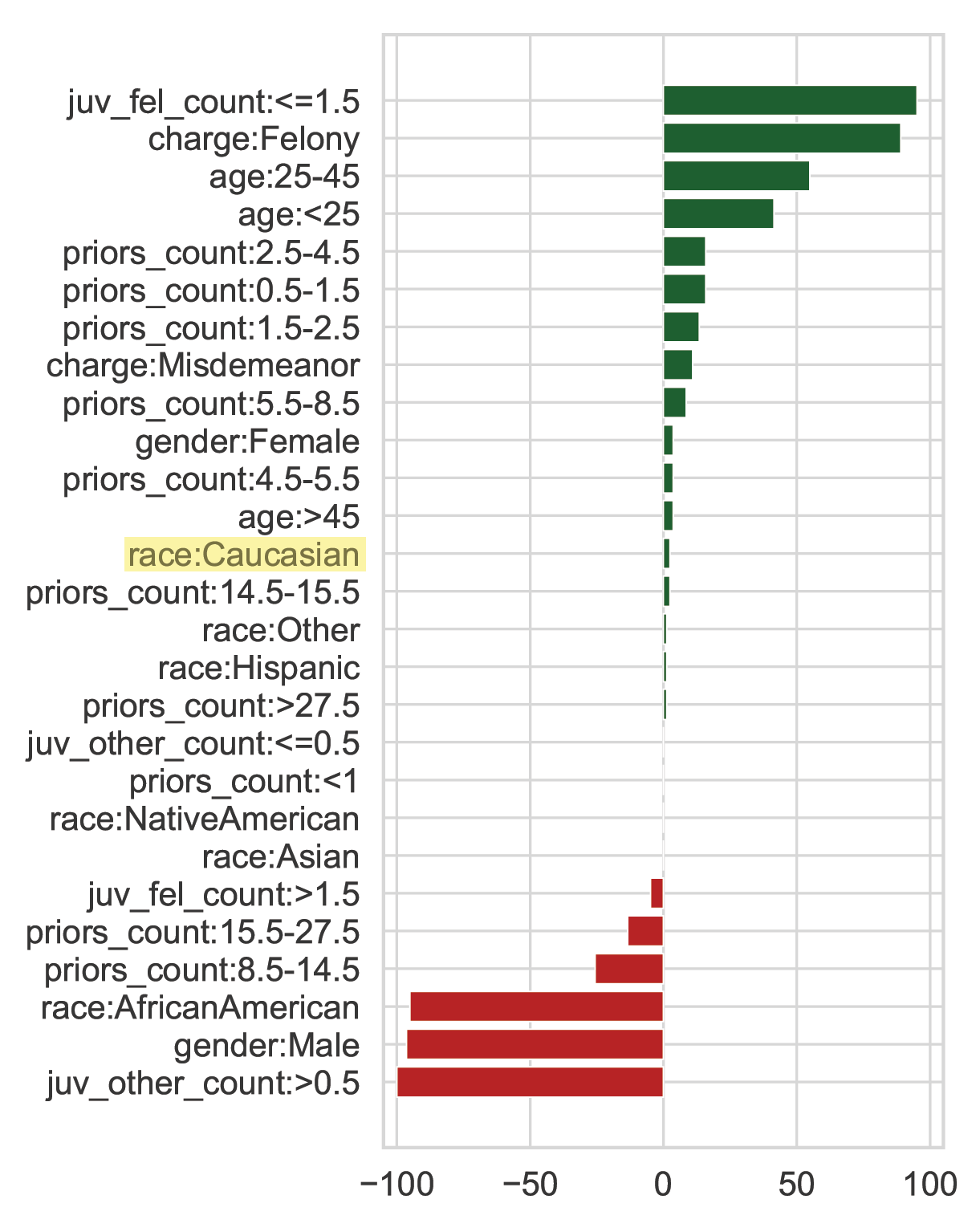}
  \includegraphics[width=0.25\textwidth]{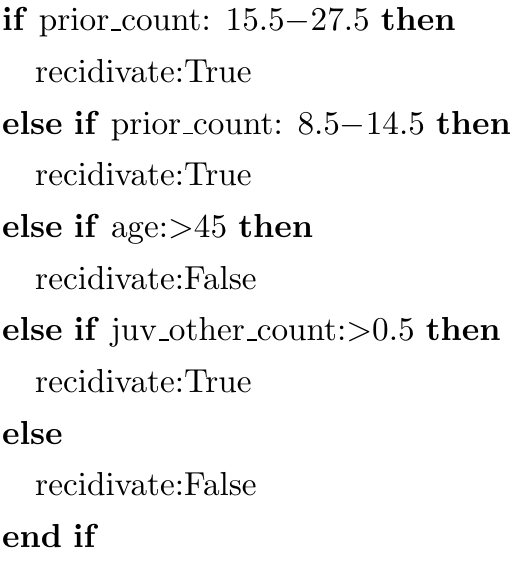}
    \caption{Relative feature dependence ranking obtained using \textsf{FairML} to audit models trained on the Adult Income dataset (upper panel) and the ProPublica Recidivism dataset (lower panel). 
    Green indicates that the feature highly contributes to a high salary rating on Adult (respectively, a low recidivism rating on ProPublica Recidivism). 
    Features that characterize the majority groups are highlighted in yellow. 
    Blackbox model (left) vs. \badml{} model (middle) and its description (right).}
    \label{fig:audit}
\end{figure*}

For the scenario (S1), we use regularization parameters with values within the following ranges $\lambda = \{0.005, 0.01\}$ and $\beta = \{0.0, 0.1, 0.2, 0.5, 0.7, 0.9\}$ for both datasets, yielding $12$ experiments per dataset. 
For each of these experiments, we enumerate $50$ models.

For the scenario (S2), we use the regularization parameters $\lambda = 0.005$ and $\beta = \{0.1, 0.3, 0.5, 0.7, 0.9\}$ for both datasets. 
For each dataset, we produce a rationalized outcome explanation of each rejected subject belonging to the minority groups (\emph{i.e.}, female subjects of Adult Income's \group{} and black subjects of ProPublica Recidivism's \group{}), and for whom the unfairness as measured in their neighbourhood is greater than $0.05$.
Overall, they are $2,634$ (respectively $285$) such subjects in the Adult Income's \group{} (respectively ProPublica Recidivism's \group{}), yielding a total of $13,170$ (respectively $1,425$) experiments for the two datasets. 
For each of these experiments, we also enumerate $50$ models and select the one (if it exists) that has the same outcome as the black-box model and the lowest unfairness. 
To compute the neighbourhood $\mathsf{neigh}(\cdot)$ of each subject, we apply the $k$-nearest neighbour algorithm with $k$ set to $10\%$ of the size of the \group{}. 
Although it would be interesting to see the effect of the size of the neighbourhood on the performance of the algorithm, we leave this aspect to be explored in future works.

\subsection{Experimental results}

\paragraph{Summary of the results.}
Across the experiments, we observed the fidelity-fairness trade-off by performing model enumeration $\badmlg{}$.
By visualizing this trade-off, we can select the threshold of the fairness condition easily. 
In some cases, the best models we found out significantly decrease the unfairness while retaining a high fidelity with the black-box model.
We have also checked that the sensitive attributes have a low impact on the surrogate models' predictions compared to the original black-box ones.
This confirms the effectiveness of the model enumeration, in the sense that we can identify a model convenient for rationalization out of the enumerated models.
In addition, by using $\badmll{}$, we can obtain models explaining the outcomes for users in the minority groups while being fairer than the black-box model. 

\paragraph{(S1) Responding to a \group{}.}
Figure~\ref{fig:globalExp-train} represents the results obtained on the scenario (S1) based on $\badmlg{}$, by depicting the changes of fidelity and unfairness of the models accompanied by the fairness regulation parameter $\beta$. 
From these results, we observe that as $\beta$ increases, both the unfairness and the fidelity of the enumerated models decrease.
In addition, as the enumerated models get more complex (\emph{i.e.}, $\lambda$ decreases), their fidelity increases. 
On both datasets, we selected the best model using the following strategy. 
First, we select models whose unfairnesses are at least two times less than the original black-box model and then among those models, and we select the one with the highest fidelity as the \emph{best model}. 
On Adult Income, the best model has a fidelity of  $0.908$ and an unfairness of $0.058$. 
On ProPublica Recidivism, the best model has a fidelity of $0.748$ and an unfairness of $0.080$.

In addition, we compared the obtained best models with the corresponding black-box models using \textsf{FairML}. 
With \textsf{FairML}, we ranked the importance of each feature to the model's decision process. 
Figure~\ref{fig:audit} shows that the best obtained surrogate models were found to be fair by \textsf{FairML}: in particular the rankings of the sensitive attributes (\emph{i.e.}, \texttt{gender:Male} for Adult Income, and \texttt{race:Caucasian} for ProPublica Recidivism) got significantly lower in the surrogate models. 
In addition, we observe a shift from the $2$nd position to the last one on Adult Income as well as a change from the $5$th rank to the $13$th rank on ProPublica Recidivism.

\begin{table}[t!]
\centering
\caption{Proportion of users of the minority group for whom there exists at least one models that agrees with the black-box models.}
\vspace{2mm}
\begin{tabular}{ccc}
\toprule
 & \textbf{Adult Income} & \textbf{ProPublica Recidism} \\
 \midrule
$\beta = 0.1$ & $92.90\%$ & $100\%$ \\
$\beta = 0.3$ & $94.95\%$ & $100\%$ \\
$\beta = 0.5$ & $97.72\%$ & $100\%$ \\
$\beta = 0.7$ & $99.16\%$ & $100\%$ \\
$\beta = 0.9$ & $100\%$ & $100\%$\\
\bottomrule
\end{tabular}
\vspace{-2mm}
\label{tab:local-perfs}
\end{table}

\paragraph{(S2) Responding to individual claim.}
Table~\ref{tab:local-perfs} as well as Figure~\ref{fig:local-adult} demonstrate the results on (S2) by using $\badmll{}$. 
From these results, we observe that as the fairness regulation parameter $\beta$ increases, the unfairness of the enumerated models decreases. 
In particular, with $\beta=0.9$, a completely fair model (unfairness = $0.0$) was found to explain the outcome for each rejected user of the minority group of Adult Income. 
For ProPublica Recidivism, the enumerated model found for each user is at least twice less unfair than the black-box models.

\paragraph{Remark.}
Our preliminary experiments in (S1) show that the fidelity of the model rationalization on the test set tends to be slightly lower than the one on the \group{}, which means that the explanation is customized specifically for the \group{}. 
Thus, if the \group{} gather additional members on which the existing explanation is applied, the explanation may not be able to rationalize as well those additional members. 
This opens the way for future research directions such as developing methods for detecting that an explanation is actually a rationalization or conversely obtaining rationalizations that are difficult to be detected.

\section{Conclusion}
\label{sec:conclusion}
In this paper, we have introduced the rationalization problem and the associated risk of fairwashing in the machine learning context and shown how it can be achieved through model explanation as well as outcome explanation. 
We have also proposed a novel algorithm called \badml{} for regularized model enumeration of rules lists, which incorporate fairness as a constraint along with the fidelity.
Experimental results obtained using real-world datasets demonstrate the feasibility of the approach.
Overall, we hope that our work will raise the awareness of the machine learning community and inspire future research towards the ethical issues raised by the possibility of rationalizing, in particular with respect to the risk of performing fairwashing in this context. 

Our framework can be extended to other interpretable models or fairness metrics as we show in the additional experiments provided in appendices.
As future works, along with ethicists and law researchers, we want to investigate the general social implications of fairwashing in order to develop a comprehensive framework to reason on this issue.
Finally, as mentioned previously we also aim at developing approaches that can estimate whether or not an explanation is likely to be a rationalization in order to be able to detect fairwashing.

\section*{Acknowledgements}
We would like to thank Daniel Alabi for the help with the CORELS code and Zachary C. Lipton for his thoughtful and actionable feedback on the paper.
S\'ebastien Gambs is supported by the Canada Research Chair program as well as by a Discovery Grant and a Discovery Accelerator Supplement Grant from NSERC.
%\clearpage

\nocite{langley00}
\bibliography{papers}
\bibliographystyle{icml2019}

\appendix
\onecolumn

\section{Generalization to other fairness metrics}
\label{app:metrics}

In this section, we present the performances of \badml{} with other fairness metrics are used. 
To control the rest of the parameters, the experiments are done only for the model rationalization algorithm \badmlg{} for a black-box model trained on the Adult Income dataset. 
We use the same black-box model trained with Random Forest, that we use in our previous experiments. 
In addition to the \emph{demographic parity} metric, we use three different fairness metrics, namely the \emph{overall accuracy equality} metric, the \emph{statistical parity} metric and the \emph{conditional procedure accuracy} metric. The definitions of all these additional fairness metrics can be found in~\cite{berk2018fairness}. For each of these scenarios, we enumerate $50$ models and use the regularization parameters $\lambda = 0.005$ and $\beta = \{0, 0.1, 0.2, 0.5, 0.7, 0.9\}$. 

\begin{figure*}[h!]
   \centering
 \includegraphics[width=0.4\textwidth]{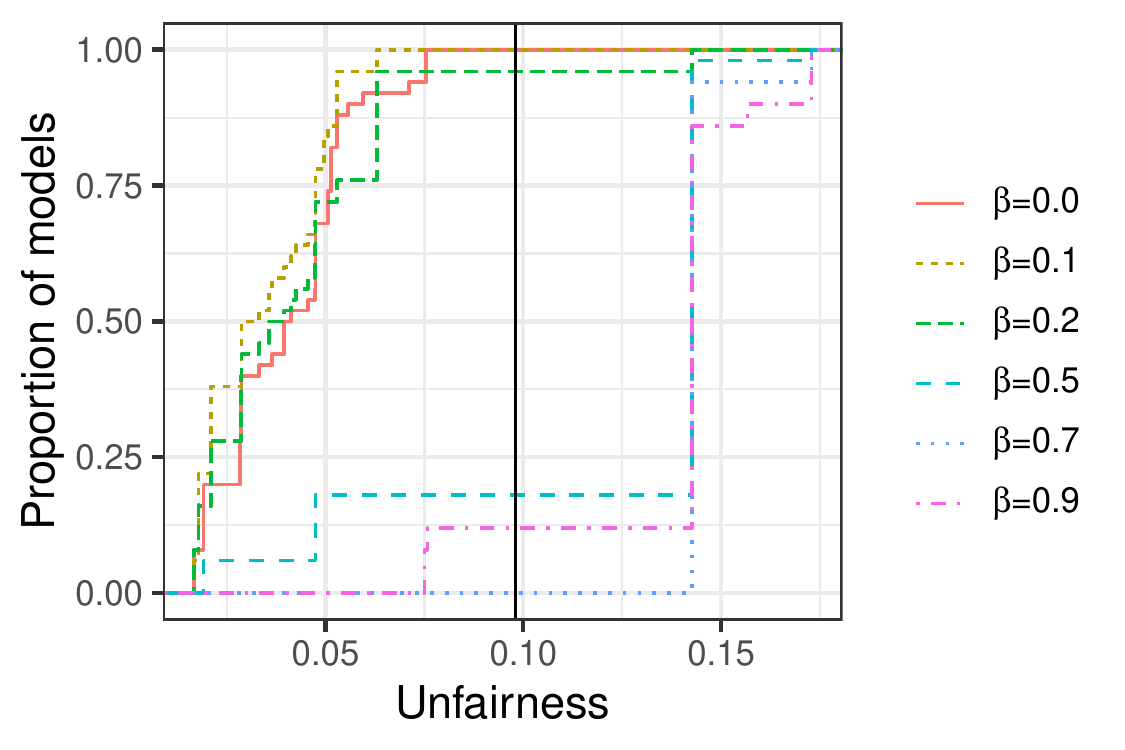}
  \includegraphics[width=0.4\textwidth]{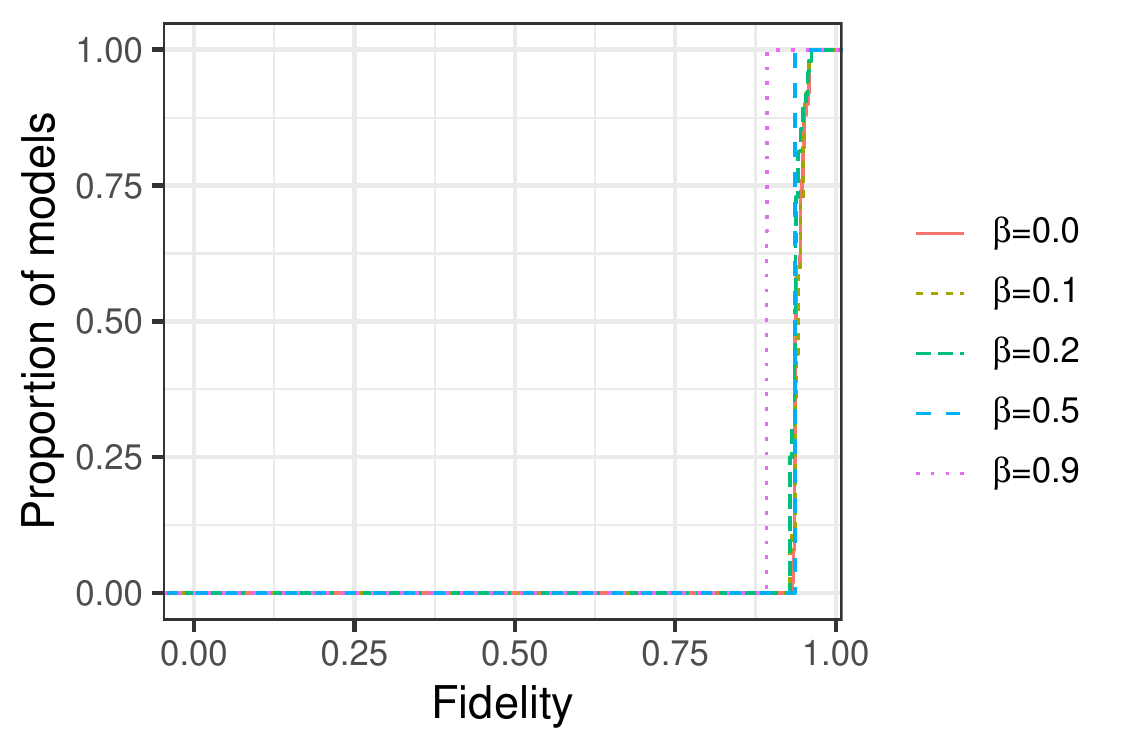}
    \caption{CDFs of the unfairness (left) and the fidelity (right) of rationalized explanation models produced by $\badmlg{}$ on the \group{}s of Adult Income. 
    Results are for the \emph{Overall Accuracy Equality} metric and the \emph{Random Forest} black-box model. 
    The vertical line on the left figure represents the unfairness of the black-box model. 
    The CDFs on the right figure are the CDFs of the fidelity of explanation models whose unfairness are less than that of the black box model.}
    \label{fig:metrics_1}
\end{figure*}

\begin{figure*}[h!]
   \centering
 \includegraphics[width=0.4\textwidth]{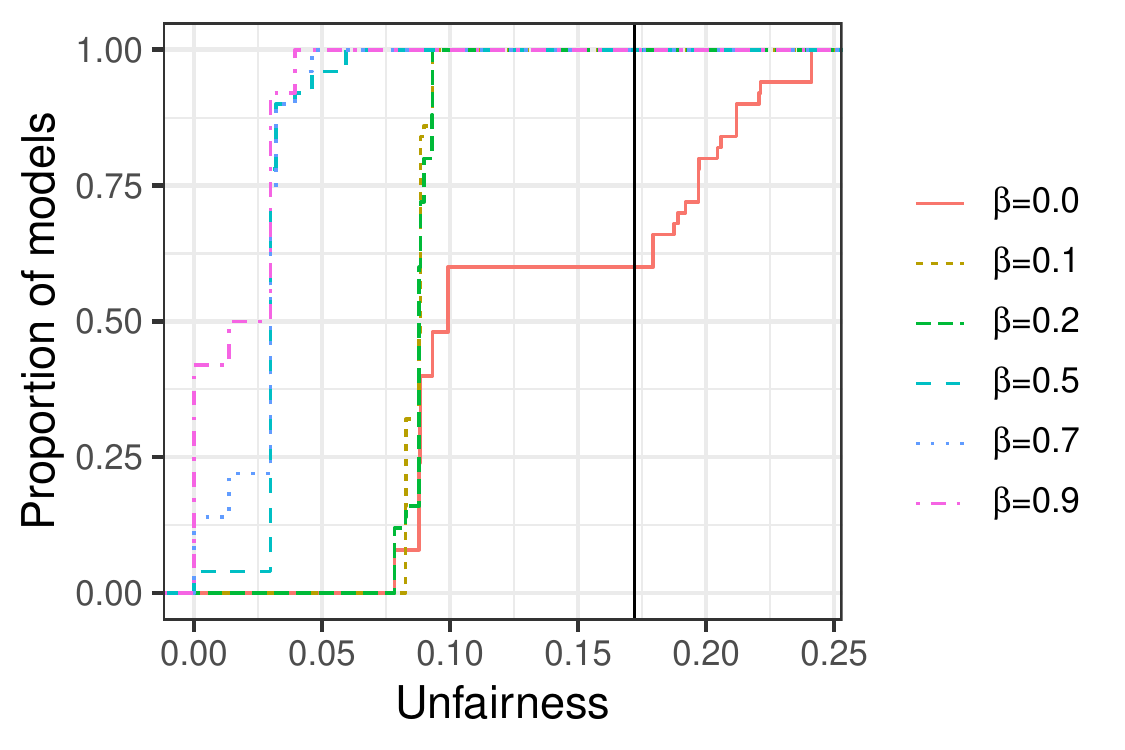}
  \includegraphics[width=0.4\textwidth]{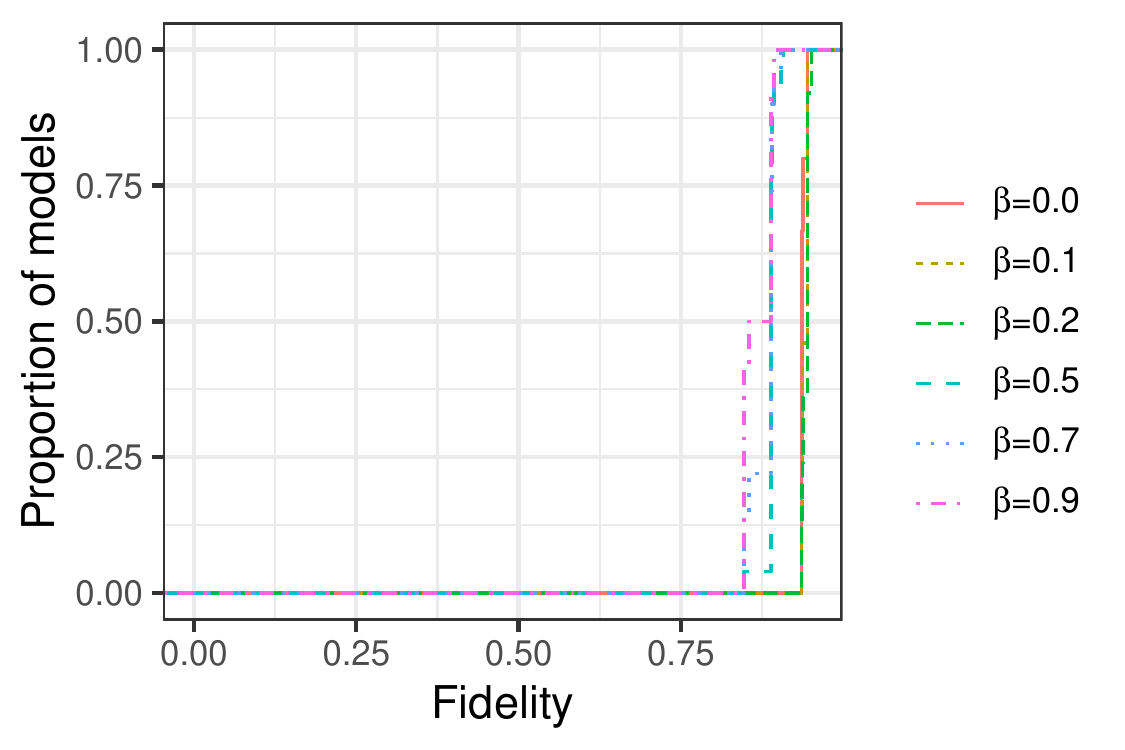}
    \caption{CDFs of the unfairness (left) and the fidelity (right) of rationalized explanation models produced by $\badmlg{}$ on the \group{}s of Adult Income. 
    Results are for the \emph{Statistical Parity} metric and the \emph{Random Forest} black-box model. 
    The vertical line on the left figure represents the unfairness of the black-box model. 
    The CDFs on the right figure are the CDFs of the fidelity of explanation models whose unfairness are less than that of the black box model.}
    \label{fig:metrics_2}
\end{figure*}

\begin{figure*}[h!]
   \centering
 \includegraphics[width=0.4\textwidth]{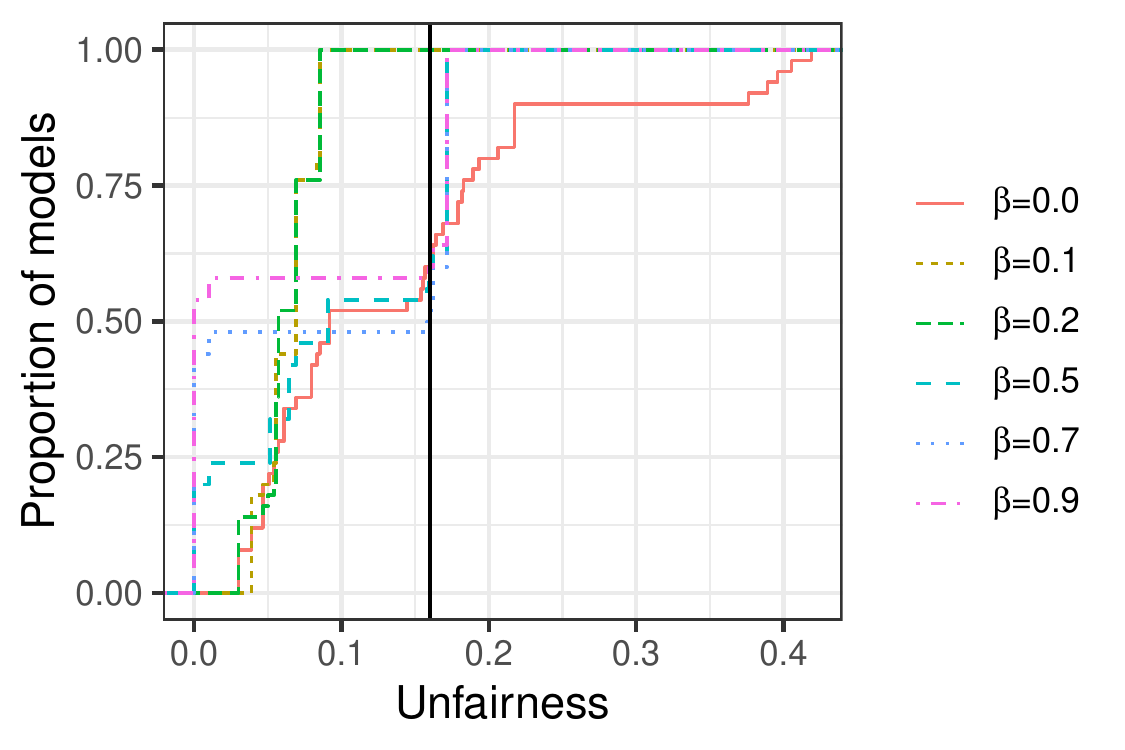}
  \includegraphics[width=0.4\textwidth]{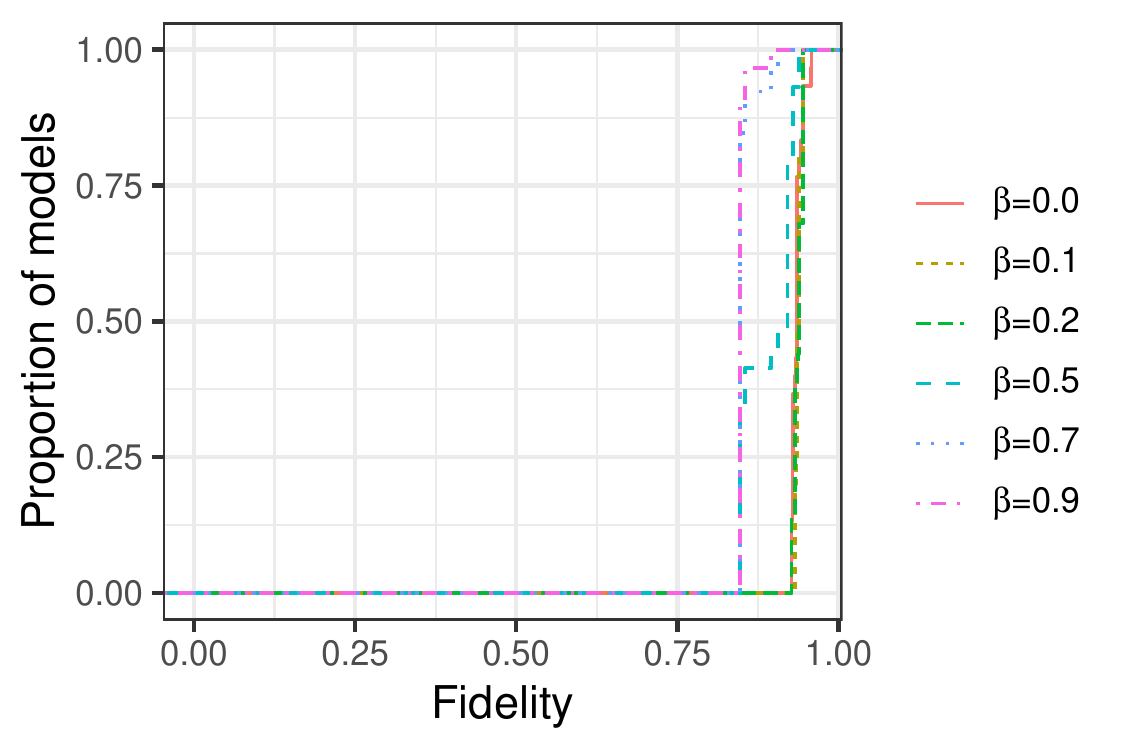}
    \caption{CDFs of the unfairness (left) and the fidelity (right) of rationalized explanation models produced by $\badmlg{}$ on the \group{}s of Adult Income. 
    Results are for the \emph{Conditional Procedure Accuracy} metric and the \emph{Random Forest} black-box model. 
    The vertical line on the left figure represents the unfairness of the black-box model. 
    The CDFs on the right figure are the CDFs of the fidelity of explanation models whose unfairness are less than that of the black box model.}
    \label{fig:metrics_3}
\end{figure*}

\begin{figure*}[h!]
   \centering
 \includegraphics[width=0.4\textwidth]{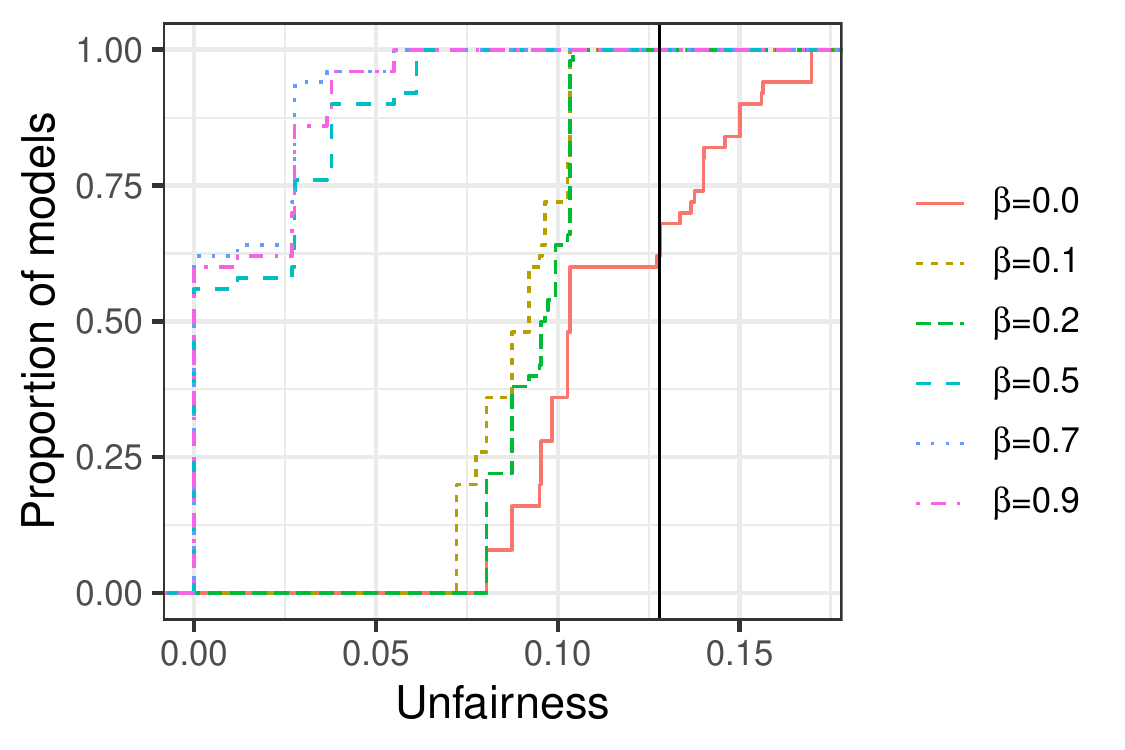}
  \includegraphics[width=0.4\textwidth]{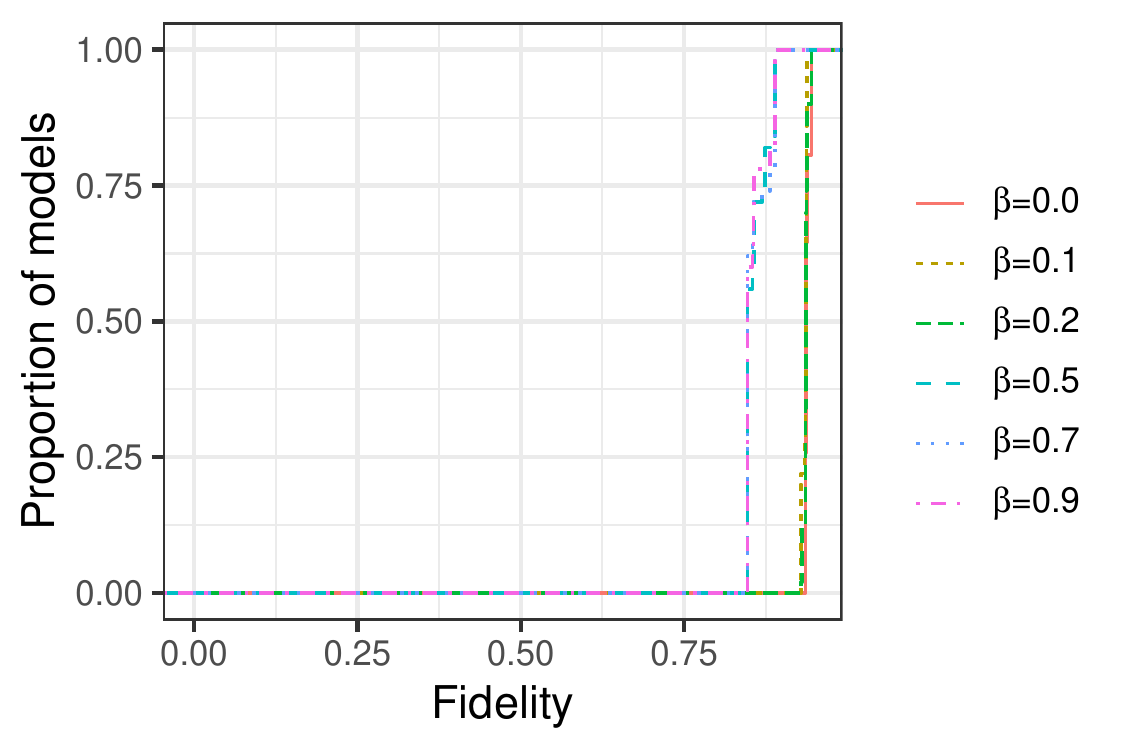}
    \caption{CDFs of the unfairness (left) and the fidelity (right) of rationalized explanation models produced by $\badmlg{}$ on the \group{}s of Adult Income. 
    Results are for the \emph{Demographic Parity} metric and the \emph{Random Forest} black-box model. The vertical line on the left figure represents the unfairness of the black-box model. The CDFs on the right figure are the CDFs of the fidelity of explanation models whose unfairness are less than that of the black box model.}
    \label{fig:metrics_5}
\end{figure*}

Figures~\ref{fig:metrics_1}, \ref{fig:metrics_2}, \ref{fig:metrics_3} and \ref{fig:metrics_5} show the performances of \badmlg{} when respectively \emph{overall accuracy equality}, \emph{statistical parity}, \emph{conditional procedure accuracy} and \emph{demographic parity} are used as fairness metrics, and the black-box model is a \emph{Random Forest} classifier. 
Overall, for all these fairness metrics, \badmlg{} can find explanation models to use for fairwashing. 
In particular, when the unfairness of the black-box model is high (\emph{i.e.,} unfairness $\geq 0.1$), a higher unfairness regularization (\emph{i.e.,} $\beta \geq 0.5$) allows to find more potential candidate models for fairwashing. 
In contrast, when the unfairness of the black-box model is low (\emph{i.e.,} unfairness $< 0.1$), a lower unfairness regularization (\emph{i.e.,} $\beta < 0.5$) enables to find more potential candidate models for fairwashing. 
In general, the smaller the regularization, the better the fidelity. 
Overall, these results confirm that the possibility of performing fairwashing is agnostic to the fairness metric considered.

\section{Generalization to other black-Box models}
\label{app:metrics}

In this section, we present the performances of \badml{} when other black-box models are used. 
To control the rest of the parameters, the experiments are done only for the model rationalization algorithm \badmlg{} with \emph{demographic parity} as fairness metrics. 
In addition to the \emph{Random Forest}, we use three different types of black-box models, namely a \emph{Support Vector Machine} (SVM) classifier, a \emph{Gradient Boosting} (XGBOOST) classifier and a \emph{Multi-Layer Perceptron} (MLP) classifier. For each of these scenarios, we enumerate $50$ models and use the regularization parameters $\lambda = 0.005$ and $\beta = \{0, 0.1, 0.2, 0.5, 0.7, 0.9\}$.

\begin{figure*}[h!]
   \centering
 \includegraphics[width=0.4\textwidth]{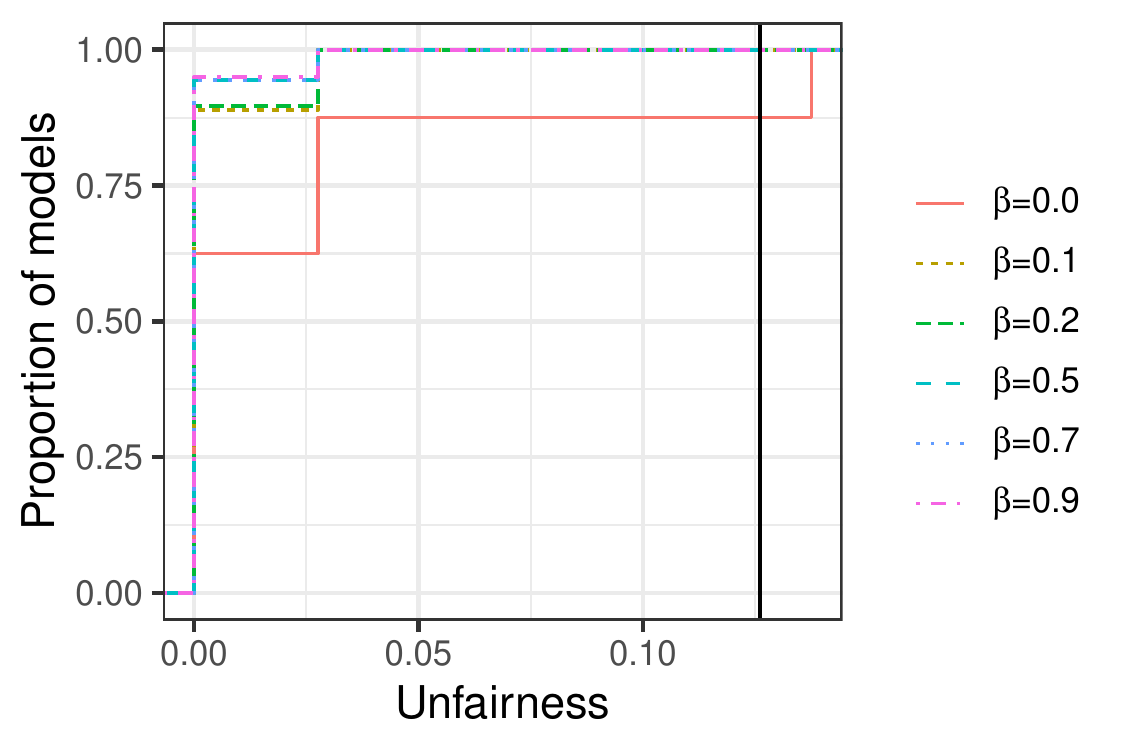}
  \includegraphics[width=0.4\textwidth]{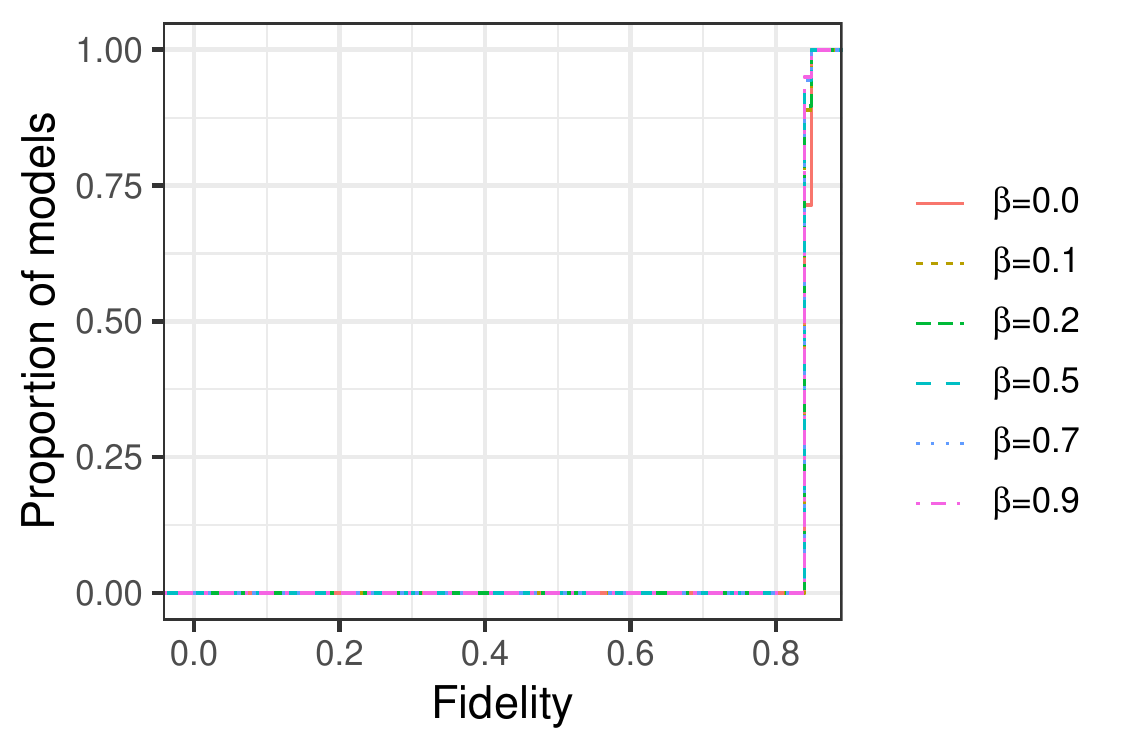}
    \caption{CDFs of the unfairness (left) and the fidelity (right) of rationalized explanation models produced by $\badmlg{}$ on the \group{}s of Adult Income. 
    Results are for the \emph{Demographic Parity} metric and the \emph{SVM} black-box model. The vertical line on the left figure represents the unfairness of the black-box model. The CDFs on the right figure are the CDFs of the fidelity of explanation models whose unfairness are less than that of the black box model.}
    \label{fig:bbox_svm}
\end{figure*}

\begin{figure*}[h!]
   \centering
 \includegraphics[width=0.4\textwidth]{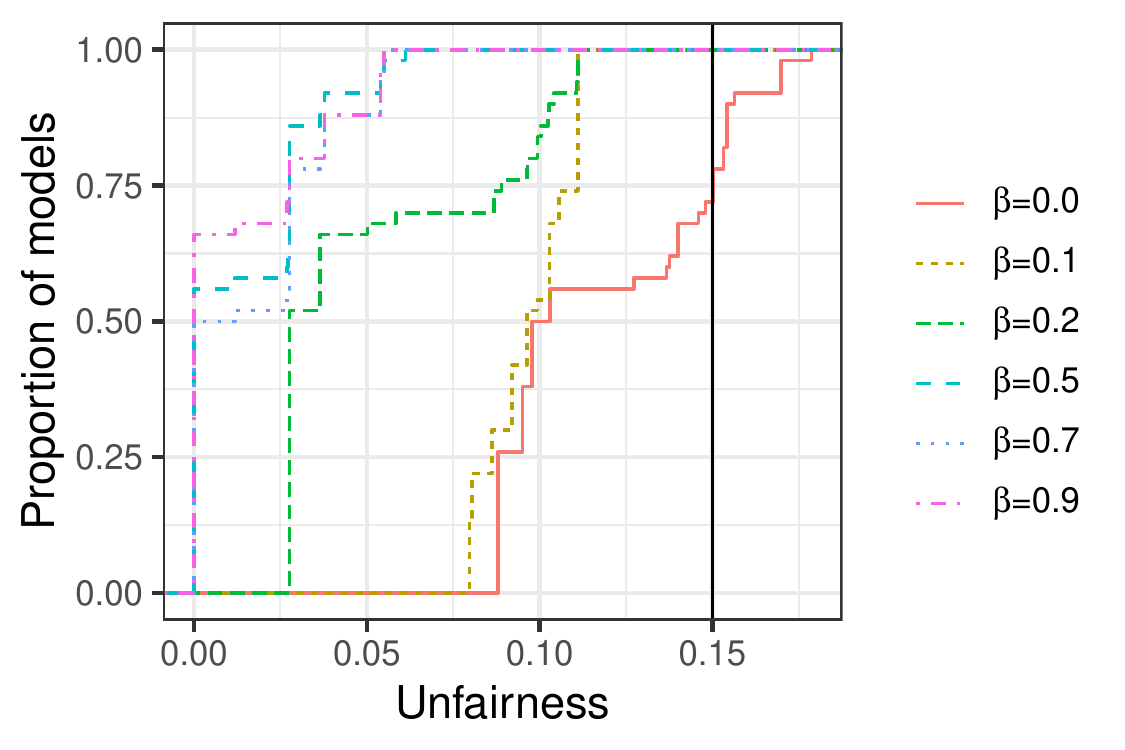}
  \includegraphics[width=0.4\textwidth]{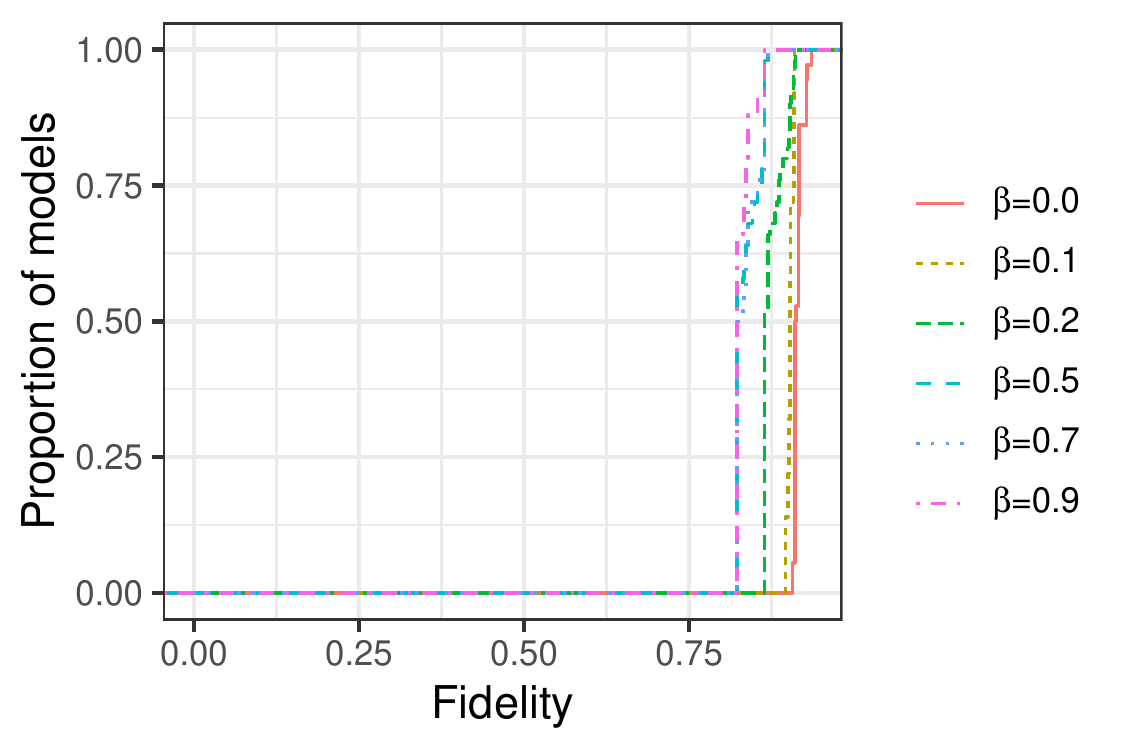}
    \caption{CDFs of the unfairness (left) and the fidelity (right) of rationalized explanation models produced by $\badmlg{}$ on the \group{}s of Adult Income. 
    Results are for the \emph{Demographic Parity} metric and the \emph{XGBOOST} black-box model. The vertical line on the left figure represents the unfairness of the black-box model. The CDFs on the right figure are the CDFs of the fidelity of explanation models whose unfairness are less than that of the black box model.}
    \label{fig:bbox_xgb}
\end{figure*}

\begin{figure*}[h!]
   \centering
 \includegraphics[width=0.4\textwidth]{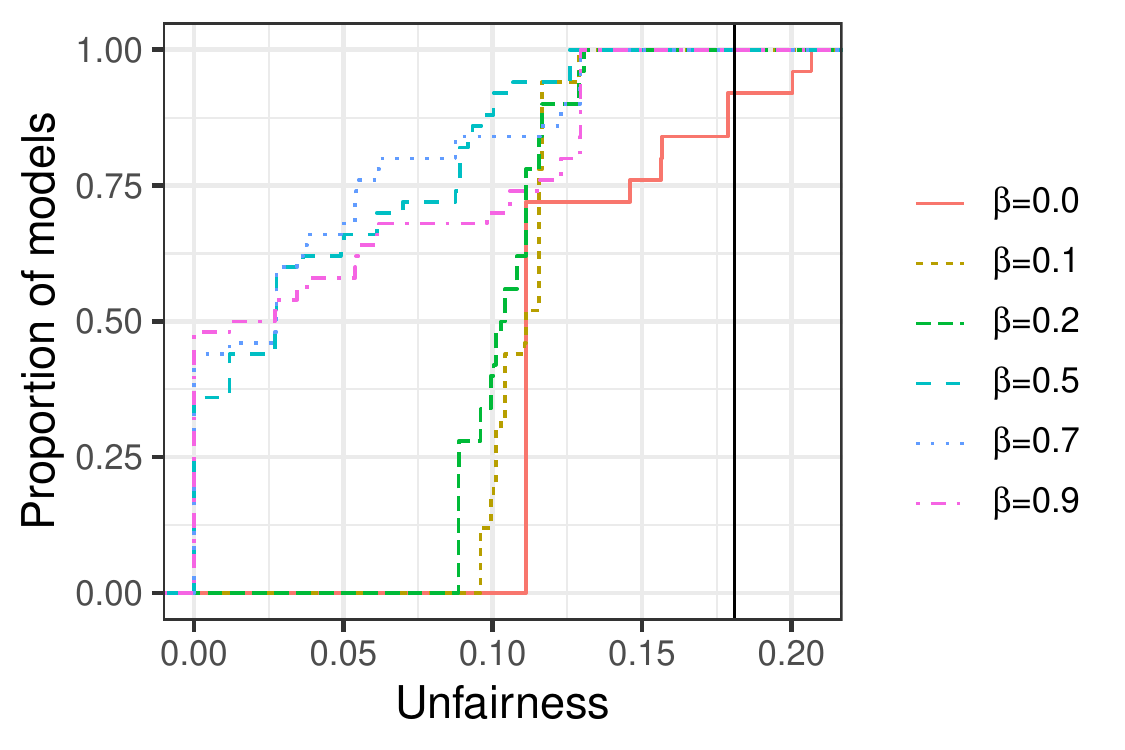}
  \includegraphics[width=0.4\textwidth]{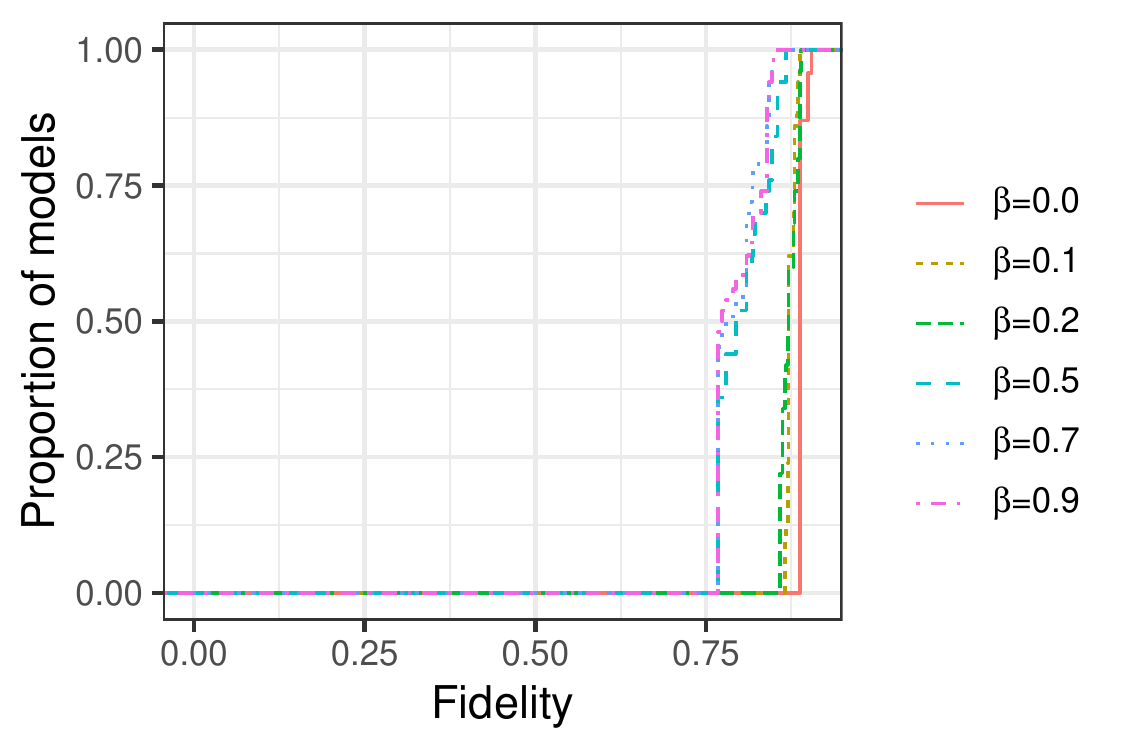}
    \caption{CDFs of the unfairness (left) and the fidelity (right) of rationalized explanation models produced by $\badmlg{}$ on the \group{}s of Adult Income. Results are for the \emph{Demographic Parity} metric and the \emph{MLP} black-box model. The vertical line on the left figure represents the unfairness of the black-box model. The CDFs on the right figure are the CDFs of the fidelity of explanation models whose unfairness are less than that of the black box model.}
    \label{fig:bbox_dnn}
\end{figure*}

Figures~\ref{fig:metrics_5}, \ref{fig:bbox_svm}, \ref{fig:bbox_xgb} and \ref{fig:bbox_dnn} show the performances of \badmlg{} when the black-box models are respectively a \emph{Random Forest} classifier, a \emph{SVM} classifier, a \emph{XGBOOST} classifier and a \emph{MLP} classifier, and the fairness metric is \emph{demographic parity}. For each type of black-box model, \badmlg{} can find explanation models to use for fairwashing. Overall, these results confirm that the possibility of performing fairwashing is also agnostic to the black-box model.

%%%%%%%%%%%%%%%%%%%%%%%%%%%%%%%%%%%%%%%%%%%%%%%%%%%%%%%%%%%%%%%%%%%%%%%%%%%%%%%
%%%%%%%%%%%%%%%%%%%%%%%%%%%%%%%%%%%%%%%%%%%%%%%%%%%%%%%%%%%%%%%%%%%%%%%%%%%%%%%

\end{document}